\documentclass[10pt]{article} % For LaTeX2e
\usepackage[preprint]{tmlr}
\usepackage{booktabs}
\usepackage{amsmath,amsfonts,bm}

\def\eqref#1{equation~\ref{#1}}
\def\1{\bm{1}}

\DeclareMathAlphabet{\mathsfit}{\encodingdefault}{\sfdefault}{m}{sl}
\SetMathAlphabet{\mathsfit}{bold}{\encodingdefault}{\sfdefault}{bx}{n}

\usepackage{graphicx}
\usepackage{hyperref}
\usepackage{cleveref}
\usepackage{url}
\usepackage{booktabs}
\usepackage{longtable}
\usepackage{pdflscape}
\usepackage{array}
\usepackage{wrapfig}

\newcommand{\modelname}{\textsc{Tabby}{}}

\newcommand{\smallmodelname}{\textsc{Tabby-Small}{}}
\usepackage{svg}
\title{%
    \parbox{\textwidth}{\centering
        \modelname: An Open Pretraining Recipe for\\
        Time Series Foundation Models
    }%
}

\author{%
    \parbox{\textwidth}{\centering
        \name
        Shifeng Xie$^{1,2}$ \quad
        Bahaeddine Abdessalem$^{1,3}$ \quad
        Zehao Xiao$^{1}$ \\[0.4em]
        \name
        Youssef Attia El Hili$^{1,4}$ \quad
        Ambroise Odonnat$^{1,5}$ \quad
        Zhiwei Dong$^{1}$ \quad
        \name
        Lei Zan$^{1}$ \quad \\[0.4em]
        Themis Palpanas$^{2}$ \quad
        Jianfeng Zhang$^{6}$ \quad
        Lujia Pan$^{6}$ \quad
        Keli Zhang$^{1}$ \quad
        Malik Tiomoko$^{1,*}$ \email malik.tiomoko@huawei.com \\[0.8em]
        \addr
        $^{1}$Huawei Noah's Ark Lab, Paris, France \\
        \addr
        $^{2}$LIPADE, Universit\'e Paris Cit\'e, Paris, France 
        \addr
        $^{3}$\'Ecole Polytechnique, France \\
        \addr
        $^{4}$Centre de Recherche en Informatique,
        Mines Paris, PSL University, France \\
        \addr
        $^{5}$IRISA, Universit\'e Rennes 2, Inria, France \\
        \addr
        $^{6}$Huawei Noah's Ark Lab, Shenzhen, China \\
        \vspace{0.3em}
        \textit{$^{*}$Corresponding author; Team Leader of the Tabby project.}\\
    }%
}
\def\month{MM}  % Insert correct month for camera-ready version
\def\year{YYYY} % Insert correct year for camera-ready version
\def\openreview{\url{https://openreview.net/forum?id=XXXX}} % Insert correct link to OpenReview for camera-ready version

\begin{document}

\maketitle
\begin{abstract}
In this report, we release \modelname, a long context probabilistic time series foundation model, together with a complete and open recipe of how it was built. \modelname{} adopts an encoder-only patch Transformer architecture and concentrates the contributions on the data and the training procedure. The pretraining corpus combines an extended real-world collection, GIFT-Eval-Pretrain+ and BLAST, with synthetic data from KernelSynth and CauKerV2, an online generator that composes temporal dynamics through randomly sampled structural causal models. Training couples a progressive convergence schedule, which yields reusable intermediate checkpoints, with a deep quantile supervision objective for intermediate layers. The resulting 145M parameter backbone supports contexts of up to 8{,}192 observations and serves forecasting, classification, and anomaly detection, while a prompt-tuning module further improves in-distribution forecasting performance with the pretrained weights frozen. \modelname{} achieves competitive zero-shot forecasting performance on GIFT-Eval and the out-of-distribution TIME benchmark, while the same pretrained backbone also supports classification on the UCR Archive and zero-shot anomaly detection on TSB-AD-U. We release training pipeline and model as open source at \href{https://github.com/huawei-noah/trustworthyAI}{\texttt{huawei-noah/trustworthyAI}}.
\end{abstract}

\section{Introduction}
Time series are widely used in healthcare \citep{TSFMHealthcare}, finance \citep{financeTSFM}, transportation \citep{TransportationTSFM}, telecommunications \citep{TelecomTSFM}, and many other domains \citep{wang2026deeptimeseriesmodels}. 
Despite their different origins, time series often share common temporal patterns, such as trends, seasonality, local dependencies, and abrupt changes \citep{survey2,surveyTSFM}. 
These shared structures make it possible to train time series foundation models (TSFMs) that transfer knowledge across datasets and domains \citep{survey3,survey1}. 
Recent models such as TimesFM-3 \citep{timesfm} and Falcon-2 \citep{Falcon} achieve strong zero-shot forecasting performance on diverse benchmarks \citep{fevbench,aksu2024gift,qiao2026s}. 

% However, recent progress has also exposed a substantial reproducibility gap\citep{patchtstFM}. 
% Constructing a large and diverse pretraining corpus requires considerable effort, while training a TSFM demands substantial computational resources. 
% Consequently, reproducing a complete pretraining process from existing papers or technical reports alone is often difficult. Moreover, seemingly minor choices such as dataset balancing, window sampling, masking strategies, normalization, learning rate scheduling, and checkpoint management can have a major influence on pretraining. Nevertheless, these practical details are often discussed only briefly in academic publications.
However, this progress remains difficult to build on. 
First, pretraining a TSFM is expensive to reproduce. Assembling a large and diverse corpus requires considerable collection effort, and the training process usually demands substantial computational resources.
Second, the process is often partially documented.
Detailed but consequential choices, such as dataset balancing, window sampling, masking strategies, normalization, learning-rate scheduling, and checkpoint management, have a major influence on the final model, yet they are often not reported in enough detail to be re-implemented.
Recent work has shown that a patch-based Transformer trained on open data with a well-tuned recipe matches specialized architectures~\citep{patchtstFM}, suggesting that these data and training decisions, rather than architectural differences, are the primary drivers of TSFM performance. 
% Documenting them fully is therefore not an engineering footnote but the core of what a technical report should deliver. 
Making these decisions explicit is therefore a
central aim of this report. 
In addition, most open recipes target zero-shot forecasting only, while practical deployments also require classification and
anomaly detection.

\begin{figure*}[t]
    \centering
    % \vspace{-5mm}
    % \includegraphics[width=\textwidth]{tabby_overview.pdf}
    \includegraphics[width=\textwidth]{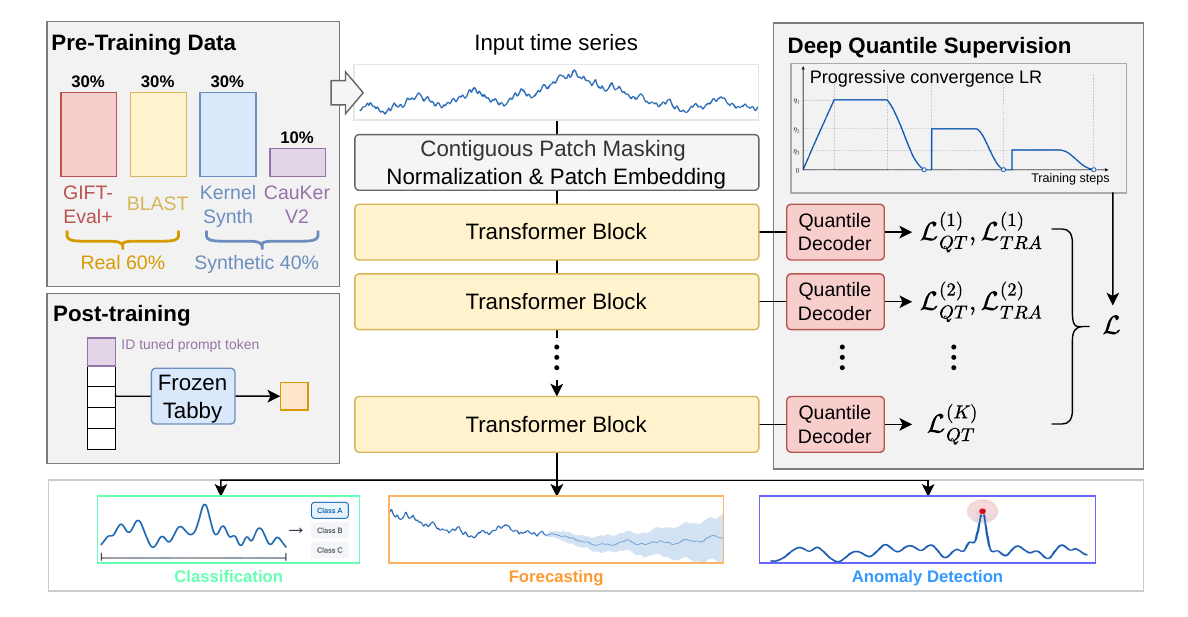}
    \vspace{-8mm}
    \caption{Overview of \modelname. The model is pretrained on real and synthetic time series using deep quantile supervision and progressive convergence learning rate schedule, and supports forecasting, classification, and anomaly detection.}
    \vspace{-2mm}
    \label{fig:tabby-overview}
\end{figure*}

In this report, we propose \modelname, a long context probabilistic TSFM, together with a complete and open account of how it was built. 
\modelname{} adopts the encoder-only patch Transformer architecture~\citep{patchtstt,patchtstFM}, and concentrates the contributions on the two factors identified above: the data and the training procedure.
On the data side, we combine a curated and extended real-world corpus (GIFT-Eval-Pretrain+ \citep{aksu2024gift} and BLAST \citep{BLAST}) with complementary synthetic data from KernelSynth~\citep{chronos} and CauKerV2.
CauKerV2 extends CauKer~\citep{cauker} from classification to forecasting, which composes ten families of temporal dynamics through randomly
sampled structural causal models and generates data online during pretraining.
On the training side, we introduce a progressive convergence learning rate schedule (PCS) that yields reusable, locally converged checkpoints throughout pretraining, and a deep quantile supervision (DQS) objective that supervises intermediate layers along the trajectory between the initial and final predictions.
In addition, we equip the backbone with a parameter-efficient prompt-tuning module for in-distribution forecasting adaptation \citep{xiao2026latent}, which further improves performance while keeping the pretrained weights frozen.
Moreover, a single pretrained backbone is then extended to three tasks: forecasting, classification, and anomaly detection. 
The pretrained model has 145.843M parameters, supports contexts of up to 8{,}192 observations, and outputs quantile forecasts. The pretraining corpus totals 229.80B GIFT-Eval-Pretrain+ \citep{aksu2024gift} and 81.92B BLAST \citep{BLAST} real-world time points and 48M synthetic samples.
An illustration of \modelname{} is provided in Figure \ref{fig:tabby-overview}.
All components, including the corpus construction recipe, the training pipeline, and the model itself, are released as open source \href{https://github.com/huawei-noah/trustworthyAI}{\texttt{huawei-noah/trustworthyAI}}.

On zero-shot forecasting, \modelname{} is competitive with recent TSFMs on
GIFT-Eval~\citep{aksu2024gift} and on the out-of-distribution TIME benchmark~\citep{qiao2026s}. Beyond forecasting, the same pretrained backbone supports classification and anomaly detection, evaluated on UCR \cite{ucr} and  TSB-AD-U~\citep{Liu2024Elephant}. 

In summary, our main contributions are:

% To address these limitations, we present this technical report and provide a detailed account of the end-to-end pretraining process, from data construction to model evaluation. Our main contributions are threefold:

\begin{itemize}
    \item \textbf{The \modelname{} model.}
    % We introduce \textsc{\modelname}, a long-context probabilistic time-series foundation model, and evaluate it on GIFT-Eval and TIME. \textsc{\modelname} achieves top-tier zero-shot forecasting performance across both benchmarks.
    We release \modelname, a long context probabilistic TSFM that serves forecasting, classification, and anomaly detection from a single pretrained backbone, with top tier zero-shot forecasting performance on GIFT-Eval \citep{aksu2024gift} and TIME \citep{qiao2026s}.
    
    \item \textbf{A feasible pretraining corpus.}
    We construct a large and diverse pretraining corpus
    that combines curated real-world time series with complementary synthetic data, most notably from CauKerV2, an online generator that extends CauKer \citep{cauker} from classification to forecasting. We release the data, the generators, and the construction recipe.
    % by combining carefully curated real-world time series with complementary synthetic data generated using KernelSynth and CauKerV2.

    \item \textbf{An open-source pretraining pipeline.}
    % We provide a complete and reproducible pipeline covering data processing, balanced sampling, synthetic data generation, distributed training, checkpoint management, and model evaluation.
    We introduce progressive convergence learning rate schedule that produces reusable intermediate checkpoints, together with deep quantile supervision objective for intermediate layers. The full pipeline, including data processing, balanced sampling, distributed training, and evaluation, is released for end-to-end reproduction.
\end{itemize}

\section{Related Work}

Time series forecasting has largely relied on task-specific models trained separately for individual datasets or application domains \citep{survey5}. Global forecasting models, such as DeepAR \citep{DeepAR}, represented an important step toward sharing statistical information across related series, while architectures such as PatchTST \citep{patchtstt} introduced patch-based representations and channel-independent modeling for efficient forecasting. Nevertheless, these models were primarily developed under dataset-specific training or downstream adaptation settings \citep{dlinear}. The emergence of time series foundation models has shifted this paradigm toward large scale pretraining followed by zero-shot forecasting on previously unseen datasets. Representative models, including TimeGPT \citep{garza2024timegpt1}, Lag-Llama \citep{lagllama}, TimesFM \citep{timesfm}, Chronos \citet{chronos}, and Moirai \citep{moirai}, employ different architectures and prediction objectives but share the goal of learning transferable temporal patterns from large and heterogeneous time series corpora.

More recent TSFMs have extended this paradigm along several dimensions. Models such as Tirex \citep{tirex}, Sundial \citep{sundial}, and PatchTST-FM \citep{patchtstFM} support probabilistic forecasting through discrete sampling, parametric distributions, generative objectives, or direct quantile prediction. Moirai-2 \citep{moirai2} and Chronos-2 \citep{chronos2} extend zero-shot forecasting to multivariate series and covariate-informed settings, while TiRex-2 \citep{tirex2} further supports long-context and streaming inference. At the frontier of model scaling, Timer-S1 \citep{timerS1} further pushes TSFMs into the billions parameter through a sparse mixture-of-experts architecture. Synthetic data have also become an important component of TSFM pretraining: ForecastPFN \citep{forecastPFN} is trained entirely on synthetic series \citep{cauker,Boris}, whereas toto-2 \citep{toto,toto2} combine real-world data with synthetic processes designed to increase temporal and cross-variable diversity. To the best of our knowledge, PatchTST-FM \citep{patchtstFM} is the model most closely related to \modelname, as both adopt a channel independent patch Transformer, masked reconstruction, long input contexts, and dense quantile prediction. \modelname{} further introduces CauKerV2 as a structurally diverse synthetic data generator, progressive convergence learning rate schedule, and deep supervision across intermediate transformer layers. In addition, we release the complete end-to-end pretraining codebase, covering data construction, synthetic data generation, distributed training, checkpoint continuation, and evaluation.

\section{Preliminary}
\label{Preliminary}
\paragraph{Univariate TSFM Task Formulation}
Let $\mathbf{x}_{1:T}=(x_1,\ldots,x_T)\in\mathbb{R}^{T}$ denote a univariate time series observed over a context window of length $T$. Given a forecasting horizon $H$, the objective is to predict the future sequence $\mathbf{y}_{1:H}=(x_{T+1},\ldots,x_{T+H})\in\mathbb{R}^{H}$. A forecasting model $f_{\theta}(\cdot)$ parameterized by $\theta$ therefore learns the mapping $\widehat{\mathbf{y}}_{1:H}=f_{\theta}(\mathbf{x}_{1:T})$. In probabilistic forecasting, the model estimates a predictive distribution for each future value rather than producing only a single point estimate.
Beyond forecasting, we also consider time series classification and anomaly detection. Given an input series $\mathbf{x}_{1:T}$, classification aims to predict a sequence-level label $y \in \{1,\ldots,C\}$. Anomaly detection instead assigns each time step an anomaly score $a_t \in \mathbb{R}_{\geq 0}$, where a larger value indicates a stronger deviation from the temporal patterns learned by the model.

% \paragraph{Patch-Based Temporal Representation}

% Following PatchTST, the input sequence is divided into patches of length $P$ and stride $S$, producing $N=\left\lfloor (T-P)/S \right\rfloor+1$ patch tokens. The $i$-th patch is defined as $\mathbf{p}_i=(x_{(i-1)S+1},\ldots,x_{(i-1)S+P})\in\mathbb{R}^{P}$ and is projected into a $d$-dimensional representation through $\mathbf{z}_i^{(0)}=\mathbf{W}_{p}\mathbf{p}_i+\mathbf{b}_{p}+\mathbf{e}_i$, where $\mathbf{e}_i$ is a learnable positional embedding. The resulting token sequence is processed by a Transformer encoder that applies self-attention across temporal patches. Compared with point-wise tokenization, patching reduces the sequence length and allows each token to summarize a meaningful local temporal pattern.

\paragraph{Quantile Forecasting Objective}

To represent predictive uncertainty, the model estimates conditional quantiles for a predefined set $\mathcal{Q}\subset(0,1)$. For each horizon step $h$ and quantile level $\tau\in\mathcal{Q}$, it predicts $\widehat{q}_{h,\tau}$ as an approximation of $q_{\tau}(y_h\mid\mathbf{x}_{1:T})=\inf\{z:\Pr(y_h\leq z\mid\mathbf{x}_{1:T})\geq\tau\}$. The model is trained using the pinball loss $\rho_{\tau}(u)=u\bigl(\tau-\mathbb{I}[u<0]\bigr)$, where $\mathbb{I}[\cdot]$ denotes the indicator function, giving the forecasting objective
\[
\mathcal{L}_{\mathrm{QT}}
=
\frac{1}{H|\mathcal{Q}|}
\sum_{h=1}^{H}
\sum_{\tau\in\mathcal{Q}}
\rho_{\tau}\!\left(y_h-\widehat{q}_{h,\tau}\right).
\]
Minimizing this objective enables the model to estimate both the central forecast and the uncertainty of future observations.

% \zx{Do we also need something about classification and anomaly detection here?}

% During pretraining, different samples may contain different numbers of
% masked target positions. Let $\mathcal{M}_i$ denote the set of masked
% and observed target positions of time series sample $i$, and let
% $n_i = |\mathcal{M}_i|$. The per-sample quantile loss at exit $\ell$ is
% \[
% \ell_i^{(\ell)}
% =
% \frac{1}{n_i |\mathcal{Q}|}
% \sum_{t\in\mathcal{M}_i}
% \sum_{\tau\in\mathcal{Q}}
% \rho_\tau
% \left(
% x_{i,t}-\hat q_{i,t,\tau}^{(\ell)}
% \right).
% \]
% The batch loss uses square-root length weighting,
% \[
% \mathcal{L}_{\mathrm{QT}}^{(\ell)}
% =
% \frac{
% \sum_i n_i^\alpha \ell_i^{(\ell)}
% }{
% \sum_i n_i^\alpha
% },
% \qquad
% \alpha=0.5.
% \]
% The trajectory-regularization MSE is aggregated using the same masking
% and length-weighting rule.
%\zx{TBD: We need a subsection or paragraph for the details of the architecture.}

\section{\modelname{} Model Architecture}
\label{sec:model-architecture}

{\modelname~is an encoder-only, non-autoregressive model that supports forecasting, classification, and anomaly detection. 
We use forecasting as the pretraining task, and extend the pretrained model to classification and anomaly detection through downstream task tuning \citep{mantis}.
}

% \modelname~is an encoder-only, non-autoregressive model that formulates forecasting as masked reconstruction within a fixed-length window. 
{During pretraining, we formulate forecasting as masked reconstruction within a fixed-length window.}
The window contains both the observed history and the masked positions to be predicted. 
% \zx{I think we can directly say: Following XXX, we use CPM to learn XXX, where the masked window contains XXX.}
After mask-aware reversible normalization (details shown in \cref{TerminalMasks}), the normalized values are divided into non-overlapping patches and mapped to token embeddings. Learned positional embeddings are then added, and the resulting tokens are processed by a stack of bidirectional transformer blocks. Since the future values are replaced by masked inputs, the encoder can attend globally without observing the prediction targets.

A shared decoder maps every patch to a set of quantiles for each time point. The quantiles are constructed from cumulative positive increments, which guarantees their monotonic ordering by design. The predictions are finally transformed back to the original scale. For multivariate inputs, the current implementation applies the same model to each channel independently. It therefore shares parameters across channels but does not explicitly model cross-channel interactions. Complete architectural equations and hyperparameters are provided in
Appendix~\ref{app:model-architecture-details}.

\section{Pretraining Data}
In this section, we present the construction and composition of the pretraining corpus used for \modelname. The corpus integrates complementary real and synthetic time series. The real-world component is primarily built upon GIFT-Eval-Pretrain and BLAST, while the synthetic component consists of KernelSynth and {our proposed} online CauKerV2 samples. 
{Both the real-world datasets and synthetic data generation pipelines will be made publicly available, supporting transparent and reproducible corpus construction.}
%A complete dataset description, including the source, domain, size and sequence-length statistics of every dataset, is provided in \cref{app:gift-eval-pretrain-plus}.

\subsection{Real-World Time Series Corpora}
The real-world component of our pretraining corpus is constructed from two primary open-source datasets: GIFT-Eval-Pretrain \citep{aksu2024gift} and BLAST \citep{BLAST}. 
%These corpora exhibit substantially different characteristics in terms of dataset composition, sampling balance, and sequence length, making them complementary for large-scale time-series pretraining.

GIFT-Eval-Pretrain consists of a heterogeneous collection of time series datasets from multiple application domains. However, its constituent datasets vary substantially in size. Directly sampling individual series from their union would cause a small number of extremely large datasets to dominate pretraining. We therefore adopt a sub-dataset-balanced sampling strategy. Let \(D\) denote the number of constituent datasets. At each sampling step, we first sample a dataset uniformly and then sample a time series from the selected dataset:
\[
d \sim \operatorname{Uniform}{1,\ldots,D},
\qquad
x \sim p_d(x),
\]
where \(p_d\) denotes the empirical distribution of the (d)-th dataset. Consequently, every constituent dataset has an expected sampling probability of \(1/D\), independently of its raw number of time series. This strategy prevents very large datasets from overwhelming the pretraining distribution.

Uniform sub-dataset sampling may, however, repeatedly expose the model to examples from extremely small datasets and thereby increase the risk of overfitting. To mitigate this issue, we merge selected small datasets from the same or closely related application domains into larger composite sampling units. 
%The balancing procedure is then applied to these consolidated units. 
This preserves the contribution of relatively small domains while avoiding disproportionate oversampling of individual small datasets.

We further expand the original GIFT-Eval-Pretrain corpus by comparing its dataset inventory with the training corpora used by Chronos and Tirex-2. We selectively incorporate complementary, high quality datasets. Importantly, the resulting corpus is not a direct union of the three training corpora: only eligible datasets that satisfy our quality, length, predictability, and benchmark-separation criteria are retained. We refer to the resulting curated and extended corpus as GIFT-Eval-Pretrain+.

Our second major real-world source is BLAST. BLAST is already constructed to provide a strongly balanced distribution of temporal patterns. We therefore stream BLAST records directly during pretraining. 
Its main limitation is the sequence length, where the available BLAST records contain at most 4,096 time steps.

The two corpora are therefore complementary. BLAST provides extensive and explicitly balanced coverage of temporal patterns, but is restricted to relatively short and fixed-width sequences. GIFT-Eval-Pretrain+ offers broader domain heterogeneity and substantially greater length diversity, although its raw constituent datasets are not intrinsically balanced and consequently require an explicit sampling strategy. Detailed dataset-level sources, filtering decisions, and corpus statistics are reported in Appendix~\ref{app:gift-eval-pretrain-plus}.

\subsection{Kernel Generated Synthetic Data}
KernelSynth was introduced together with the Chronos family and has since been widely adopted for pretraining of TSFMs. It samples time series from Gaussian processes defined by randomly composed kernels.
A major limitation of KernelSynth is its computational cost. Constructing and sampling from a Gaussian-process covariance matrix becomes increasingly expensive as the sequence length grows, with the practical generation cost increasing at least quadratically with the target context length. Moreover, the available generation pipeline is CPU-bound.

To improve efficiency, we generate each sequence on a temporal grid that is four times coarser than the target resolution and subsequently upsample it to the desired context length through interpolation. For a target length \(T\), KernelSynth therefore operates on approximately \(T/4\) points. Under quadratic scaling, this reduces the main generation cost by approximately
\(
\frac{(T/4)^2}{T^2}=\frac{1}{16},
\)
while the interpolation step introduces only linear overhead. Since individual synthetic sequences and output shards can be generated independently, we additionally distribute the generation process across multiple parallel CPU tasks. This combination of coarse-grid generation, interpolation, and task-level parallelism makes large-scale long-context KernelSynth pretraining substantially more practical.

\subsection{CauKerV2}
% \zx{Better to have one start sentence for CauKerV2. Why CauKerV2 better than KS? what it can bring for the synthetic data? For example: Although Kernal Synth is widely used it is (limitations). To XXX (advantages, such as improve the diversity of synthetic data), we further propose CauKerV2 as another important source of the pretraining data.}
Although KernelSynth provides useful smooth and temporal dependent series, its kernel-based construction captures only a limited range of temporal patterns. Moreover, the high computational cost of Gaussian process sampling makes it difficult to generate sufficiently diverse samples online during pretraining. To complement KernelSynth and reduce overfitting to a fixed synthetic corpus, we introduce CauKerV2 as an additional source of pretraining data.
CauKerV2 extends CauKer \citep{cauker} by using a broader collection of primitive time series generators. Rather than relying solely on Gaussian processes, CauKerV2 spans ten families of temporal dynamics commonly observed across diverse domains. Table~\ref{tab:caukerv2-generators} summarizes the generators used to construct the synthetic time series.

\begin{table}[t] \centering \small 
\caption{Primitive generator families used by CauKerV2. The listed application
domains are representative rather than exhaustive.}
\begin{tabular}{p{4.8cm}p{4.5cm}p{6.0cm}} 
\toprule 
\textbf{Generator family} & \textbf{Temporal characteristics} & \textbf{Representative domains} \\ 
\midrule Compositional Kernel Gaussian Process & Smoothness and local correlations & Weather, energy, environmental sensing \\ 
Trend Seasonality Decomposition & Trends with additive or multiplicative seasonality & Retail, traffic, electricity demand \\ 
Autoregressive Integrated Moving Average & Linear dependence, integration, and seasonal autoregression & Economics, demand, climate \\ 
Regime-Switching Ornstein--Uhlenbeck Stochastic Differential Equation & Mean reversion, stochastic fluctuations, and regime transitions & Finance, physical systems \\ 
Piecewise Level and Change-Point Process & Abrupt or smooth level shifts and piecewise-stationary behavior & Industrial monitoring, operations \\ 
Spike and Event Process & Spikes, bursts, shocks, plateaus, and recovery patterns & Web traffic, energy, healthcare \\ 
Analytic Waveform Mixture & Sawtooth, square, and triangular periodic signals & Sensors, machinery, electronics \\ 
Fractional Brownian Motion and Fractional Gaussian Noise & Long-range dependence and scale-dependent stochasticity & Network traffic, geophysics, finance \\ 
Generalized Autoregressive Conditional Heteroskedasticity & Volatility clustering, heavy tails, and asymmetric shocks & Financial and energy markets \\ 
Chaotic & Lorenz, Mackey--Glass, NARMA, and multi-scale oscillatory dynamics & Physical systems, audio, biomedical signals \\ 
\bottomrule \end{tabular} 
% \caption{Primitive generator families used by CauKerV2. The representative domains are illustrative rather than exhaustive.} 

\label{tab:caukerv2-generators} \end{table}

For each synthetic sample, CauKerV2 draws a set of primitive series and combines them through a randomly sampled structural causal model (SCM \cite{scm,tabpfn,TimePFN}). Each primitive series is assigned to a root node, while every non-root node is generated from its parents as
\[
z_j(t)
=
\phi_j\left(
\sum_{i\in\operatorname{Pa}(j)}
w_{ji}z_i(t)
\right)
+
\epsilon_j(t),
\]
where $\operatorname{Pa}(j)$ denotes the parents of node $j$, $w_{ji}$ is a randomly sampled edge weight, and $\epsilon_j(t)$ is additive noise. The nonlinear function $\phi_j$ is sampled from linear, ReLU, leaky-ReLU, sigmoid, sine, and modulo transformations.

The resulting random SCM combines heterogeneous primitives into higher diverse temporal processes with nonlinear interactions and varying causal graphs. 
%In our implementation, each sample contains nine primitive roots and nine observed nodes, with at most two parents per observed node. 
CauKerV2 generates these samples online during pretraining, providing a virtually unlimited synthetic corpus. Although the SCM produces multivariate series, \modelname{} treats each output channel as an independent univariate training example.

% \zx{One closing paragraph for synthetic data? like the final paragraph of 5.1}

KernelSynth and CauKerV2 play complementary roles in the synthetic pretraining corpus. KernelSynth provides statistically structured, smooth, and quasi-periodic series, whereas CauKerV2 broadens the
synthetic distribution to include discontinuities, events, regime changes, long-range dependence, heteroskedasticity, and chaotic dynamics. Importantly, CauKerV2 generates samples online throughout pretraining. This continually refreshed data stream reduces repeated exposure to the same
synthetic samples and consequently mitigates overfitting to a finite synthetic dataset. Together, the two generators provide both structured regularity and broad temporal diversity.

\section{Pretraining Methodology}
% \zx{Do we also need a summary paragraph for section 6? like that of section 5.}
This section presents the main components of our pretraining methodology. First, mask sampling combines contiguous internal masks with short terminal masks to balance reconstruction and forecasting while maintaining training stability. Second, our progressive convergence learning rate schedule produces well-annealed intermediate checkpoints and enables flexible intervention during pretraining. Third, deep quantile supervision directly supervises intermediate layers and regularizes their predictions toward progressive refinement across model depth. Together, these components improve the stability, flexibility, and effectiveness of large scale TSFM pretraining.

\subsection{Hybrid Mask Sampling}
\label{TerminalMasks}
We train \modelname~through masked patch reconstruction, following the general masked modeling paradigm of PatchTST-FM and the Contiguous Patch Masking (CPM) strategy of TiRex \citep{tirex}. Given a sequence of patch tokens, we mask contiguous spans and optimize the model only on the masked positions. For an internal span, which can exploit observations on both sides, we mask several contiguous patches; for a terminal span, where the model must extrapolate solely from the preceding context, we restrict its length to at most two patches. We empirically find that longer terminal masks cause substantial training instability, as further examined in Section~\ref{sec:mask_ablation}. We hypothesize that internal reconstruction provides a better-conditioned learning signal, resembling the bidirectional masked reconstruction objective used by BERT \citep{devlin2019bertpretrainingdeepbidirectional}, whereas long terminal masks impose a considerably harder extrapolation task. A complete theoretical explanation of this phenomenon remains an important direction for future investigation.

\subsection{Progressive Convergence Learning Rate Schedule}

% Pretraining a TSFM is computationally expensive, making reliable intermediate checkpoints essential for evaluating the model and adjusting hyperparameters or data mixture ratios during training. Conventional cosine schedules are less suitable for this purpose because their learning rates are coupled to a predefined training horizon, and intermediate checkpoints are obtained before complete annealing. 
{Pretraining a TSFM is computationally expensive, making it desirable to obtain well-annealed intermediate checkpoints that can be reliably evaluated throughout training, for example when adjusting hyperparameters or data mixture ratios. Conventional cosine schedules are less suited to this setting because the learning-rate trajectory is tied to a predefined training horizon, leaving intermediate checkpoints only partially annealed.}
Inspired by Warmup--Stable--Decay (WSD \cite{wsd}) and WSD-S \citep{wsd-s}, we introduce progressive convergence learning rate schedule of a single warm-up phase followed by repeated stable--decay (SD) stages. For stage $k$, the stable learning rate is
\(
\eta_k=\eta_1\,2^{-(k-1)},
\)
such that each stage operates at half the stable learning rate of its predecessor and ends with a decay phase that produces a locally converged checkpoint. These checkpoints provide explicit intervention points from which training can be evaluated, reconfigured, or continued. As shown in Section~\ref{sec:lr-convergence}, this schedule provides greater training flexibility while achieving performance comparable to or better than a conventional cosine schedule.

\subsection{Deep Quantile Supervision}

Recent studies have identified substantial architectural and representational redundancy in TSFMs, showing that intermediate layers can often be removed with only limited performance degradation \citep{wiliński2025exploringrepresentationsinterventionstime,bao2026universalredundanciestimeseries}. Meanwhile, the functional roles of individual layers remain insufficiently understood. Inspired by deeply supervised learning and the path-interpolation perspective of flow matching \citep{lipman2023flowmatchinggenerativemodeling,lee2014deeplysupervisednets,liu2022flowstraightfastlearning,sundial}, we propose deep quantile supervision. 
Specifically, we attach a shared quantile decoder to a set of decoded
exits $\mathcal{E}$, where $M$ denotes the number of Transformer blocks.
Let $\hat{\mathbf q}^{(\ell)}$ denote the quantile predictions produced
after block $\ell$. The terminal exit is optimized using the original
quantile objective, while only the intermediate exits receive auxiliary
deep supervision:
\[
\mathcal{L}_{\mathrm{DS}}
=
\sum_{\ell\in\mathcal{E}\setminus\{0,M\}}
\left(\frac{\ell}{M}\right)^\gamma
\mathcal{L}_{\mathrm{QT}}^{(\ell)}.
\]

We additionally regularize each intermediate prediction toward a linear
trajectory between the detached initial and terminal predictions:
\[
\widetilde{\mathbf q}^{(\ell)}
=
\left(1-\frac{\ell}{M}\right)
\operatorname{sg}\!\left(\hat{\mathbf q}^{(0)}\right)
+
\frac{\ell}{M}
\operatorname{sg}\!\left(\hat{\mathbf q}^{(M)}\right),
\]
\[
\mathcal{L}_{\mathrm{TRA}}
=
\sum_{\ell\in\mathcal{E}\setminus\{0,M\}}
\operatorname{MSE}_{\alpha}
\left(
\hat{\mathbf q}^{(\ell)},
\widetilde{\mathbf q}^{(\ell)}
\right),
\]
where $\operatorname{sg}(\cdot)$ denotes the stop-gradient operation and
$\operatorname{MSE}_{\alpha}$ denotes the masked, square-root
length-weighted MSE defined using the same aggregation as
$\mathcal{L}_{\mathrm{QT}}$.

The complete pretraining objective is
\[
\mathcal{L}
=
\mathcal{L}_{\mathrm{QT}}^{(M)}
+
\lambda_{\mathrm{DS}}\mathcal{L}_{\mathrm{DS}}
+
\lambda_{\mathrm{TRA}}\mathcal{L}_{\mathrm{TRA}}.
\]

For the \modelname{} model,
\[
M=20,\qquad
\mathcal{E}=\{0,5,10,15,20\},\qquad
\lambda_{\mathrm{DS}}=0.5,\qquad
\lambda_{\mathrm{TRA}}=0.1,\qquad
\gamma=1.
\]
Here, exit $0$ and exit $20$ serve as the trajectory endpoints, while
$\{5,10,15\}$ are the auxiliary deeply supervised exits.

\section{Prompt-Based In-Distribution Tuning}

We also design a lightweight, fully reversible form of supervised adaptation for \modelname: prompt tuning of the frozen model. When training data from the target distribution (e.g., GIFT-Eval) is available, a small prompt module can be trained on it to further improve performance without modifying any pretrained parameter.
% Omitting the prompt recovers the original model exactly.

\paragraph{Module.} To introduce soft prompts into the pretrained model, we prepend $M{=}160$ learnable tokens $P(x)\in\mathbb{R}^{M\times d}$ to the patch-embedding sequence, where $d$ is the latent dimension of the patch embeddings. 
All prompt tokens are assigned position index $0$, so they act as position-agnostic conditioning signals: the history keeps its full context budget and its original positional encoding. 
The forecasting process then becomes
\[
\widehat{\mathbf{y}}_{1:H} \;=\; f_\theta\big([\,P(x)\,;\,\mathbf{x}_{1:T} \,]\big),
\]
where $f_\theta$ is the frozen backbone, $\mathbf{x}_{1:T}$ is the patch embeddings of the input window $x$, and $\hat q_{1:H}$ denotes the predicted quantiles over the forecast horizon $H$.

Specifically, the prompt combines a shared component with a gated, input-adaptive one,
\[
P(x) \;=\; P_{\text{shared}} \;+\; \sigma(g)\,P_{\text{adaptive}}(\mathbf{x}_{1:T}),
\]
where $P_{\text{shared}}\in\mathbb{R}^{M\times d}$ is learned during training. $g$ is also a learnable scalar, adjusted by the sigmoid function $\sigma(\cdot)$. 
The adaptive component is generated from a compact descriptor $s(x)\in\mathbb{R}^{10}$ of the normalized input window, collecting scale-invariant statistics of its location, spread, trend, roughness, autocorrelation and length. 
A two-layer MLP maps $s(x)$ to the coefficients of a low-rank token basis
% \[
% P_{\text{adaptive}}(x) \;=\; \sum_{k=1}^{r} w_k\big(s(x)\big)\,B_k,
% \qquad w(s)=\mathrm{MLP}(s)\in\mathbb{R}^{r},\;\; B_k\in\mathbb{R}^{M\times d},
% \]
with $r{=}4$, following~\cite{xiao2026latent}. Since $s(x)$ is
low-dimensional, this factorization keeps the input-dependent part of the module small while still tailoring the prompt to each series.

\paragraph{Training.} During the in-distribution tuning, only the prompt module are updated. The backbone model are always frozen. 
The module is trained with the pretraining quantile (pinball) loss on the training portion of the target data under time-based splits, so no evaluation data is ever seen. 
A single module is trained once across all training datasets and utilized for every evaluation task. Model selection uses validation loss on held-out windows of the training split. Training costs a few GPU-hours on a single device and can be repeated for any checkpoint of the backbone. 
In inference the module is strictly optional.

\section{Evaluation}
\subsection{Forecasting}
\paragraph{Benchmarks.}
We evaluate \modelname{} on two complementary forecasting benchmarks: GIFT-Eval, and TIME. GIFT-Eval \citep{aksu2024gift} contains 23 datasets covering more than 144,000 time series across seven domains and ten sampling frequencies, with forecasting horizons ranging from short to long term. TIME \citep{qiao2026s} is a task-centric benchmark containing 50 recently collected datasets and 98 forecasting tasks. It emphasizes data freshness, data integrity, providing a stringent evaluation with a substantially reduced risk of pretraining-data contamination.

We use the terms in-distribution and out-of-distribution to describe the relationship between a benchmark and the pretraining corpus. Since our real-world pretraining corpus is largely derived from GIFT-Eval-Pretrain+ and BLAST, and \modelname{} is prompt-tuned on GIFT-Eval training set, we consider GIFT-Eval as an in-distribution benchmark. In contrast, TIME consists primarily of newly collected data sources released after many existing TSFMs were developed and therefore serves as our main out-of-distribution benchmark.
% fev-bench provides an broader evaluation setting because it combines established forecasting collections with newly curated, covariate-rich tasks.

\paragraph{Evaluation Metrics.}
We primarily evaluate point and probabilistic forecasting performance using Mean Absolute Scaled Error (MASE) and Continuous Ranked Probability Score (CRPS), respectively. For a historical context $\mathbf{x}_{1:T}$, a forecast horizon $H$, and seasonal period $m$, MASE and CRPS are defined as
\begin{equation}
\operatorname{MASE}
=
\frac{
\frac{1}{H}
\sum*{h=1}^{H}
\left|y_{T+h}-\widehat{y}*{T+h}\right|
}{
\frac{1}{T-m}
\sum*{t=m+1}^{T}
\left|x_t-x_{t-m}\right|
},
\qquad
\operatorname{CRPS}
\approx
\frac{2}{H|\mathcal{Q}|}
\sum_{h=1}^{H}
\sum_{\tau\in\mathcal{Q}}
\rho_{\tau}
\left(
y_{T+h}-\widehat{q}*{T+h}^{(\tau)}
\right)
\end{equation}
where the median forecast is used as the point prediction $\widehat{y}_{T+h}$. MASE is scale independent and measures forecasting error relative to an in-sample seasonal-naive predictor. Values below one indicate an improvement over the seasonal-naive reference.
Because \modelname{} directly predicts quantiles, we compute CRPS using its discrete quantile approximation, where $\mathcal{Q}$ denotes the quantile grid required by each benchmark and $\rho*{\tau}$ is the pinball loss. 
% For fev-bench, we additionally report the official Scaled Quantile Loss, together with the corresponding win rates and skill scores, to ensure comparability with its standard evaluation protocol.

\paragraph{Main Results.}

\begin{figure*}[t]
    \centering
    \includegraphics[width=\textwidth]{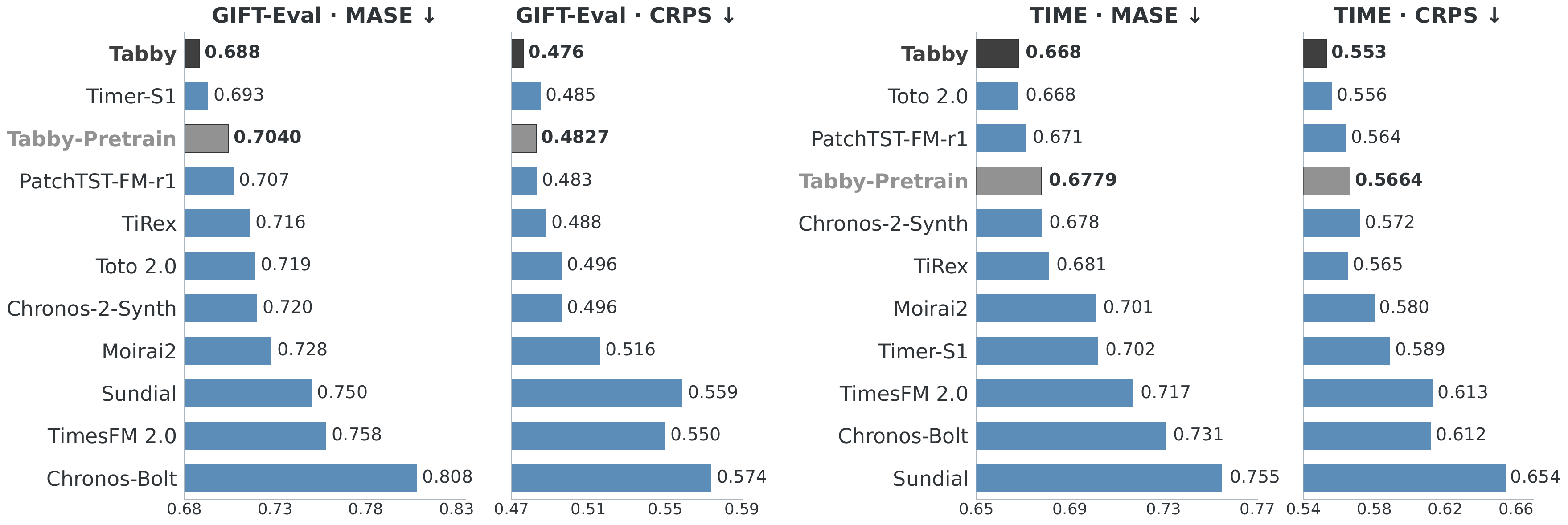}
    \caption{
        Forecasting performance on GIFT-Eval (ID) and TIME (OOD).
        Models are ordered by MASE within each benchmark.
        % \zx{The method name and numbers are too small}
        % Lower values indicate better performance.
        % \modelname{} is highlighted in dark gray, while
        % \pretrainmodelname{} is shown in light gray.
    }
    \label{fig:main_forecasting_results}
\end{figure*}

Figure~\ref{fig:main_forecasting_results} compares \modelname{} with recent state-of-the-art time-series foundation models. On the in-distribution GIFT-Eval benchmark, \modelname{} outperforms Sundial, TimesFM~2.5, and Chronos-Bolt on both metrics. On the out-of-distribution TIME benchmark, \modelname{} consistently outperforming Moirai-2.0, Sundial, TimesFM~2.5, and Chronos-Bolt. These results indicate that the balanced pretraining corpus and the proposed training methodology provide competitive forecasting performance.

\subsection{Representation} 
\label{sec:ucr_representation} 
\paragraph{Benchmarks and Evaluation Metrics} We investigate whether the representations learned by \modelname{} transfer to time series classification. We evaluate the frozen encoder on all 128 datasets in the UCR Archive \citep{ucr}. We refer to this setting as zero-shot representation learning: the pretrained encoder receives no classification-specific training and is never updated using UCR labels, although a supervised classifier is fitted on the official training split of each downstream dataset. For each dataset, missing observations are interpolated and every time series is linearly resized to a fixed length (1,024 time steps). The resulting sequence is right-aligned within the model context window and passed through the frozen Transformer encoder. The forecasting quantile head is discarded. Let \(\mathbf{H}_{i}^{(\ell)}\in\mathbb{R}^{N\times d}\) denote the patch representations of sample \(i\) after Transformer layer \(\ell\), and let \(\mathcal{V}_{i}\) denote the set of non-padding patches. We first obtain one representation from each layer using padding-aware mean pooling: 
\[ \mathbf{r}_{i}^{(\ell)} = \frac{1}{|\mathcal{V}_{i}|} \sum_{j\in\mathcal{V}_{i}} \mathbf{H}_{i,j}^{(\ell)}. \] 
Representations from different depths may encode complementary temporal information. Instead of selecting a single intermediate layer, we concatenate the pooled representations from all \(M\) Transformer layers: 
\[ \mathbf{z}_{i} = \mathbf{r}_{i}^{(1)} \mathbin{\|}\mathbf{r}_{i}^{(2)} \mathbin{\|}\cdots\mathbin{\|} \mathbf{r}_{i}^{(M)} \in\mathbb{R}^{Md}, \]
where \(\mathbin{\|}\) denotes feature concatenation. This layer-fusion strategy uses every depth uniformly and requires neither dataset-specific layer selection nor validation on the test split. For each UCR dataset, a Random Forest is trained on the fused representations and labels from its official training split. The fitted classifier is then evaluated once on the official test split. The pretrained encoder remains frozen throughout the entire procedure, and neither test labels nor test accuracy are used for model or representation selection. 

\paragraph{Main Results} 
\label{sec:ucr_results} 
Table~\ref{tab:ucr_results} reports the unweighted mean test accuracy over all 128 UCR datasets. Without any classification-specific pretraining or encoder adaptation, \modelname{} achieves an average accuracy of \(0.7999\). It outperforms Catch22+\citep{catch22}, TabPFN\citep{tabpfn}, TabICL\citep{tabicl}, MOMENT\citep{moment}, TiVIT-H\citep{tivit}, NuTime\citep{lin2024nutimenumericallymultiscaledembedding}, and Mantis\citep{mantis}, showing that \modelname{} pretraining produces representations that transfer effectively to classification.

\begin{table}[t] \centering \small 
\caption{ Mean classification accuracy over all 128 datasets in the UCR Archive.} \label{tab:ucr_results} 
\resizebox{\textwidth}{!}{ \begin{tabular}{lccccccccccc} 
\toprule & Catch22+ & TabPFN & TabICL & MOMENT & TiVIT-H  & NuTime & MantisV1 & \modelname \\
\midrule UCR & 0.7969 & 0.7806 & 0.7707 & 0.7789 & 0.7943 & 0.7732 & 0.7816 & \textbf{0.7999} \\ 
\bottomrule \end{tabular} } \end{table}

\subsection{Anomaly Detection}
\label{sec:tsbad_anomaly}
We next ask whether \modelname{}'s representations also transfer to anomaly
detection, a task where many pretrained models have been reported to struggle
against simple statistical detectors~\citep{Liu2024Elephant}. We evaluate on
TSB-AD-U~\citep{Liu2024Elephant}, a curated univariate benchmark, and report
VUS-PR~\citep{paparrizos2022vus}, which the benchmark identifies as its most
reliable measure.

As in Section~\ref{sec:ucr_representation}, we discard the quantile head and
score in representation space. Each series is tiled into non-overlapping
context windows, and the patch embeddings from the final transformer block are
pooled across all windows into \(\mathbf Z=[\mathbf z_1,\ldots,\mathbf z_{n_p}]^\top
\in\mathbb R^{n_p\times d}.\). We
then fit a Gaussian to the series' own embeddings and score each patch by its
Mahalanobis distance to it,
% \[
% a_i=\sqrt{(\mathbf{p}_i-\bm{\mu})^{\top}\bm{\Sigma}_{\lambda}^{-1}(\mathbf{p}_i-\bm{\mu})},
% \qquad
% \bm{\Sigma}_{\lambda}=(1-\lambda)\bm{\Sigma}+\lambda\tfrac{\operatorname{tr}(\bm{\Sigma})}{d}\mathbf{I},
% \]
\[
a_i
=
\sqrt{
\max\left\{
(\mathbf z_i-\boldsymbol\mu)^\top
\boldsymbol\Sigma_\lambda^{+}
(\mathbf z_i-\boldsymbol\mu),
0
\right\}
},
\]
with shrinkage \(\lambda=0.1\) and a pseudo-inverse for stability. Scores are
broadcast to the timesteps of their patch and min--max scaled. The model is
frozen and no labels are used; the only per-series estimation is
\(\bm{\mu},\bm{\Sigma}\), which assumes anomalies are a minority of the series.
The distance metric was selected on the 48-series tuning split and all reported
numbers are on the held-out evaluation split. Further details are given in
Appendix~\ref{app:tsbad}.

\begin{table}[t]
\centering
% \small
\caption{
Mean VUS-PR over the TSB-AD-U evaluation split. Baseline numbers are taken from
the TSB-AD leaderboard.}

\label{tab:tsbad_results}
\resizebox{\textwidth}{!}{
\begin{tabular}{lcccccccc}
\toprule
&
Lag-Llama
& Chronos
& TimesFM
& MOMENT (ZS)
& MOMENT (FT)
& xLSTMAD
& TSPulse
& \modelname \\
\midrule
TSB-AD-U
& 0.27
& 0.27
& 0.30
& 0.38
& 0.39
& 0.40
& 0.48
& \textbf{0.43} \\
\bottomrule
\end{tabular}
}
\end{table}

\paragraph{Main Results}
\label{sec:tsbad_results}
Table~\ref{tab:tsbad_results} reports the mean VUS-PR over the 350 evaluation
series. \modelname{} attains 0.4282, ahead of the general-purpose
foundation models evaluated zero-shot on this benchmark, including MOMENT
\citep{moment}, TimesFM \citep{timesfm}, Chronos \citep{chronos} and Lag-Llama
\citep{lagllama}. It also exceeds MOMENT in its fine-tuned form and
xLSTMAD \citep{xlstmad}, a purpose-built reconstruction detector trained on
each target series, indicating that the gain comes from the representation
rather than from adaptation to the target data. Stronger results come from methods that add
task-specific specialisation \modelname{} does not use: TSPulse
\citep{ekambaram2026tspulse} pairs a masked-reconstruction model with a
detection-specific scoring pipeline, while CHARM \citep{pastrana2026giving} and
Time-RCD \citep{lan2025foundationmodelszeroshottime} rely respectively on
anomaly-free reference data from the target series and on pretraining with
anomaly labels. \modelname{} estimates its reference distribution from the full
series it scores, without labels or clean reference data.

\section{Analysis}

\paragraph{Reference Model for Analysis.}
Due to the computational cost of repeatedly pretraining the full \modelname{} model, the following analyses are conducted using a smaller reference model, denoted as \smallmodelname. It contains 40.291M parameters and consists of 12 Transformer layers with a hidden dimension of 512. The model is pretrained for 110{,}000 optimization steps using eight GPUs and a global batch size of 512. Unless otherwise specified, all ablation studies use the same architecture, data mixture, masking strategy, and optimization
configuration, changing only the component under investigation. The complete architecture and pretraining configuration of \smallmodelname{} are provided in Appendix~\ref{app:tabby-small-configuration}.

\subsection{Analysis of Progressive Convergence Schedule}
\label{sec:lr-convergence}

\begin{figure*}[t]
    \centering
    \includegraphics[width=\textwidth]{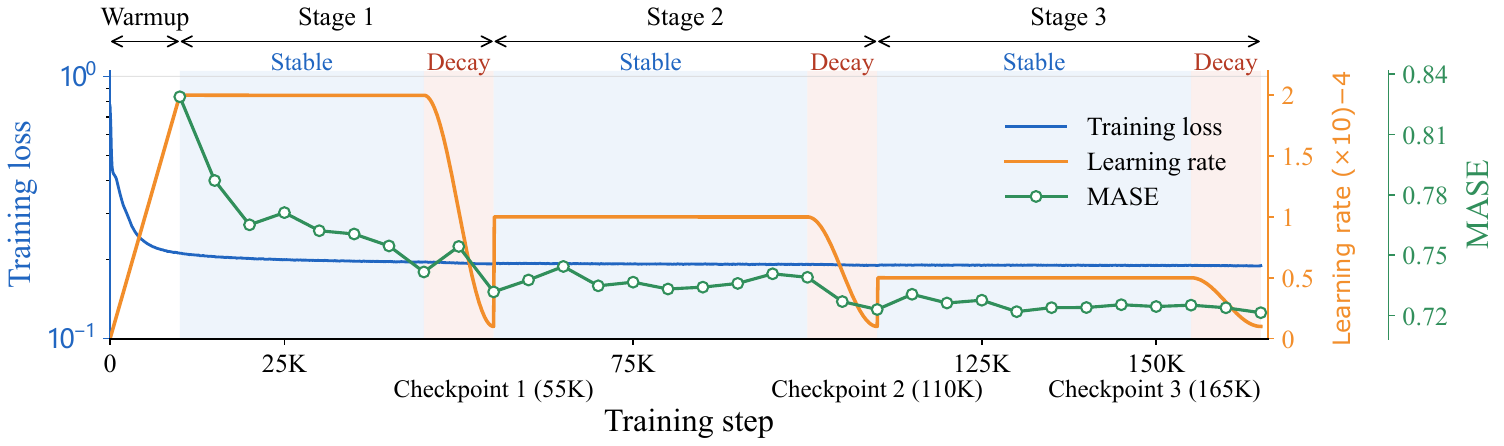}
    \caption{
    MASE is evaluated on Gift-Eval benchmark. Each stable phase is followed by a
    decay phase, producing a locally converged intermediate checkpoint.
    }
    \label{fig:multistage-lr-convergence}
\end{figure*}

\paragraph{Stage-End Convergence.}
Large-scale TSFM pretraining requires intermediate checkpoints that are
not merely arbitrary checkpoint, but locally annealed states suitable
for reliable evaluation and subsequent continuation.
In the pretraining of \modelname, we begin with a warm up. The schedule contains three stable--decay stages. Figure~\ref{fig:multistage-lr-convergence} shows the complete training trajectory over optimization steps.
The training loss remains smooth across the entire trajectory. In particular, the learning rate changes discontinuously at 55K and 110K steps: it is raised from the decay to the stable rate of the next stage. Nevertheless, neither transition produces a visible loss spike or a sustained increase in optimization noise. The model therefore remains stable when training is resumed from a stage-end checkpoint.
Within each stable phase, the loss decreases gradually and eventually approaches a relatively flat region. The subsequent decay reduces the update magnitude and produces an additional decrease before the curve reaches a new plateau. Every decay phase yields a competitive checkpoint, while later stages continue to improve upon the preceding one.

% These results support the main design goal of the proposed schedule: pretraining does not need to rely on a single checkpoint obtained only at the end of a predetermined training horizon. Instead, each decay phase acts as a controlled convergence window and produces a reusable, empirically locally converged checkpoint. Such checkpoints can be
% evaluated or used as starting points for changes to the data mixture and training configuration without restarting pretraining.

\begin{wraptable}{r}{0.35\textwidth}
    \centering
    \vspace{-0.2cm}
    \small
    \setlength{\tabcolsep}{3pt}
    \renewcommand{\arraystretch}{1.1}
    \caption{
    Learning rate ablation.
    }
    \label{tab:lr-ablation}
    \begin{tabular}{@{}lcc@{}}
        \toprule
        Schedule & MASE & CRPS \\
        \midrule
        WSD-S
        & 0.7551
        & 0.5165 \\
        Cosine
        & 0.7456
        & 0.5112 \\
        PCS
        & \textbf{0.7368}
        & \textbf{0.5038} \\
        \bottomrule
    \end{tabular}
    \vspace{-3mm}
\end{wraptable}

\paragraph{Matched-Budget Scheduler Comparison.}
We further compare the proposed schedule with Cosine \citep{patchtstFM} and WSD-S \citep{wen2024understandingwarmupstabledecaylearningrates} learning rate schedule using \smallmodelname{} at the same training budget of 110K steps. Both variants use the same pretraining pipeline, changing only the learning-rate schedule. Table~\ref{tab:lr-ablation} reports the MASE and CRPS on GIFT-Eval benchmark.
In our situation, the PCS reduces MASE and CRPS, the proposed schedule improves both point and probabilistic forecasting scores over Cosine and WSD-S schedule.

% Together with the training dynamics above, these results support
% the practical value of repeated stable--decay stages.
% Each decay phase provides an empirically locally converged
% checkpoint for evaluation or subsequent training, while the
% matched-step comparison shows that this flexibility does not
% come at the expense of forecasting performance.
% Such checkpoints provide explicit opportunities to adjust the
% data mixture or training configuration without restarting
% pretraining from scratch.

\begin{wraptable}{r}{0.53\textwidth}
    \vspace{-0.7cm}
    \centering
    \small
    \setlength{\tabcolsep}{3.2pt}
    \renewcommand{\arraystretch}{1.08}
    \caption{
        Sensitivity to restart peak learning rates.
        $G_0$ is the validation loss obtained with
        $\eta_{\mathrm{peak}}=10^{-5}$; the remaining columns report
        the relative increase over $G_0$.
    }
    \label{tab:restart-peak-sensitivity}
    \resizebox{\linewidth}{!}{
    \begin{tabular}{llrrrr}
        \toprule
        Model & Start
        & $G_0$
        & $\Delta_{5\times10^{-5}}$
        & $\Delta_{10^{-4}}$
        & $\Delta_{2\times10^{-4}}$ \\
        \midrule
        Small & 55K
        & 0.09618 & +0.42\% & +1.15\% & +2.49\% \\
        Small & 110K
        & 0.09464 & +0.63\% & +1.58\% & +3.22\% \\
        \midrule
        Main & 55K
        & 0.09031 & +0.44\% & +1.25\% & +2.65\% \\
        Main & 110K
        & 0.08856 & +0.60\% & +1.53\% & +3.60\% \\
        Main & 165K
        & 0.08793 & +0.69\% & +1.57\% & +3.63\% \\
        \bottomrule
    \end{tabular}
    \vspace{-5mm}
    }
\end{wraptable}

\paragraph{Sensitivity to aggressive restarts.}
A further motivation for decreasing the peak learning rate across
stable--decay stages is that later checkpoints become less tolerant of
aggressive restarts. We test this hypothesis using short restart
probes from the 55K and 110K checkpoints of \smallmodelname{} and from
the 55K, 110K, and 165K checkpoints of the \modelname { }model.
For each checkpoint, we restore both the model and AdamW states and
continue training with
$\eta_{\mathrm{peak}}\in
\{10^{-5},2.5\times10^{-5},5\times10^{-5},10^{-4},2\times10^{-4}\}$.
Each branch consists of 1K constant rate updates followed by a
500 step cosine cooldown to $10^{-5}$ and is evaluated on the same
fixed pretraining-distribution validation probe. Table
\ref{tab:restart-peak-sensitivity} reports the relative increase in
weighted validation loss with respect to the
$\eta_{\mathrm{peak}}=10^{-5}$ continuation. 
% Across both model scales, later stage-end checkpoints exhibit steeper
% learning-rate response curves, with most of this change emerging by
% 110K. 
This reduced tolerance to large restart updates motivates
progressively attenuating the peak learning rate rather than repeatedly
restoring its initial value. The increasingly steep response
curves show that restoring an early-stage peak becomes progressively
more disruptive. This observation provides a mechanistic motivation
for using smaller restart peaks at later stages. Complete settings and
results are provided in Appendix~\ref{app:restart-peak-sensitivity}.

\subsection{Analysis of Deep Quantile Supervision}
\label{sec:analysis-qds}

% We analyze Quantile Deep Supervision (QDS) from two complementary
% perspectives: forecasting performance and the functional utility
% of individual Transformer blocks.

\begin{wraptable}{r}{0.35\textwidth}
    \vspace{-0.7cm}
    \centering
    \small
    \setlength{\tabcolsep}{3pt}
    \renewcommand{\arraystretch}{1.1}
    \caption{GIFT-Eval DQS ablation.}
    \label{tab:qds-ablation}
    \begin{tabular}{@{}lcc@{}}
        \toprule
        Setting & MASE & CRPS \\
        \midrule
        Without DQS & 0.7476 & 0.5117 \\
        With DQS & \textbf{0.7368} & \textbf{0.5038} \\
        \bottomrule
    \end{tabular}
    \vspace{-3mm}
\end{wraptable}

\paragraph{Forecasting Performance Comparison.}
We compare \smallmodelname{} with and without
deep quantile supervision at the same training budget
of 110K steps. Both variants use the same architecture, data
mixture, masking strategy, and learning rate schedule.
Table~\ref{tab:qds-ablation} reports the MASE and weighted
quantile loss over the same 97 GIFT-Eval evaluation entries.
Removing DQS increases MASE and CRPS by \(1.47\%\) and
\(1.57\%\), respectively, supporting the benefit of
deep quantile supervision for both point and
probabilistic forecasting on GIFT-Eval.

\begin{wraptable}{r}{0.6\textwidth}
    \vspace{-0.7cm}
    \centering
    \small
    \setlength{\tabcolsep}{6pt}
    \renewcommand{\arraystretch}{1.1}
    \caption{
        Functional ablation of intermediate blocks on GIFT-Eval.
        Positive values indicate degradation.
    }
    \label{tab:qds-middle-layer-ablation}
    \begin{tabular}{@{}lcccc@{}}
        \toprule
        & \multicolumn{2}{c}{$\Delta$MASE (\%)}
        & \multicolumn{2}{c}{$\Delta$CRPS (\%)} \\
        \cmidrule(lr){2-3}
        \cmidrule(lr){4-5}
        Skipped layer
        & w/ DQS & w/o DQS
        & w/ DQS & w/o DQS \\
        \midrule
        6 & +1.13 & +0.11 & +0.91 & -0.05 \\
        7 & +1.52 & +0.09 & +1.07 & -0.14 \\
        8 & +1.63 & +0.93 & +1.45 & +0.92 \\
        \bottomrule
    \end{tabular}
\end{wraptable}
\paragraph{Functional Utility of Intermediate Blocks.} We assess the functional utility of intermediate
blocks by bypassing one block at a time during inference, without fine-tuning.
Table~\ref{tab:qds-middle-layer-ablation} presents the results for layers 6 to 8, using the zero-based indexing of
Appendix~\ref{app:qds-analysis}.
Each relative change is measured against the corresponding checkpoint with all blocks retained.

Without DQS, bypassing layers 6 and 7 has little effect on
aggregate MASE and slightly improves CRPS.
With DQS, bypassing either block increases MASE by
\(1.13\%\)--\(1.52\%\).
Bypassing layer 8 also causes larger degradation with DQS
than without it.
These results suggest that DQS encourages useful
transformations at intermediate depth and reduces local
functional redundancy in the tested checkpoints.
This interpretation concerns aggregate sensitivity to
single-block removal; it does not imply that blocks with
small aggregate effects are irrelevant to every dataset.
Full experimental settings and layer-wise results are
provided in Appendix~\ref{app:qds-analysis}.

\paragraph{DQS Improve the Local Update Direction}
\label{sec:qds-local-direction}
The model-level ablation shows that DQS improves the final forecasting performance, but it does not identify why. One possible explanation is purely scalar: adding the auxiliary loss may only increase the gradient norm and therefore induce a larger effective step. We test the stronger hypothesis that DQS changes the update direction itself and makes it more favorable to the held-out final forecasting objective.

\begin{wraptable}{r}{0.6\textwidth}
    \vspace{-0.7cm}
    \centering
    \small
    \setlength{\tabcolsep}{5pt}
    \renewcommand{\arraystretch}{1.1}
    \caption{Optimizer-aware one-step DQS intervention. Entries are the
    macro-average relative reduction in final validation pinball loss; a
    positive value favors the DQS branch B. Parentheses give 95\% confidence
    intervals from paired original-dataset cluster bootstrap.}
    \label{tab:qds-local-direction-main}
    \begin{tabular}{@{}ccc@{}}
        \toprule
        Radius $\rho$ & B over A (\%) & B over norm-matched M (\%) \\
        \midrule
        $0.25$ & $0.00712$ $[0.00112,\,0.01499]$
               & $0.00663$ $[0.00124,\,0.01373]$ \\
        $0.50$ & $0.01399$ $[0.00253,\,0.02873]$
               & $0.01296$ $[0.00251,\,0.02594]$ \\
        $1.00$ & $\mathbf{0.02698}$ $[0.00470,\,0.05446]$
               & $\mathbf{0.02477}$ $[0.00485,\,0.04930]$ \\
        \bottomrule
        \vspace{-5mm}
    \end{tabular}
\end{wraptable}
We conduct an optimizer-aware, one-step intervention at the
\smallmodelname{} checkpoint. From the same parameters, AdamW state, update
batch, and mask realization, we construct three branches: final supervision
only (A), a norm-matched rescaling of the final-loss gradient (M), and final
supervision plus DQS (B). M has the same raw gradient norm as B but preserves
the direction of the final-loss gradient. After applying the same stored
AdamW moments and clipping rule, we additionally normalize the three actual
parameter displacements to the same radius. We then evaluate the final
pinball loss on temporally disjoint held-out windows that are not used to
construct the update. The experiment uses 23 sufficiently long GIFT-Eval
configurations from 16 original dataset groups, with four paired trials per
configuration and radius multipliers $\rho\in\{0.25,0.5,1\}$. Full details
are provided in Appendix~\ref{app:qds-local-direction}.

The norm-matched comparison is the key control. The DQS gradient increases
the raw gradient norm by approximately $43.2\%$, but M reproduces this norm
without adding the DQS direction; A and M are statistically indistinguishable,
whereas B remains better than M after the actual parameter displacement is
matched. Moreover, the cosine similarity between the final and DQS gradients
is $0.717$, showing that DQS contributes a substantial non-parallel component.
After AdamW preconditioning, B improves update--validation-gradient alignment
over M. These results provide direct local
evidence that DQS supplies a more favorable update direction beyond its
effect on gradient magnitude. They do not, by themselves, establish that
every DQS step decreases held-out loss: at $\rho=1$, the loss reduction of B
relative to making no update has a positive point estimate but a confidence
interval that narrowly includes zero. The claim is therefore local and
comparative, complementing rather than replacing the end-to-end DQS
ablation.

\subsection{Analysis of CauKerV2}
\label{sec:analysis-caukerv2}

\begin{wraptable}{r}{0.35\textwidth}
    \vspace{-0.9cm}
    \centering
    \small
    \setlength{\tabcolsep}{3pt}
    \renewcommand{\arraystretch}{1.1}
    \caption{CauKerV2 ablation.}
    \label{tab:caukerv2-ablation}
    \begin{tabular}{@{}lcc@{}}
        \toprule
        Setting & MASE & CRPS \\
        \midrule
        Without CauKerV2
        & 0.7536
        & 0.5109 \\
        With CauKerV2
        & \textbf{0.7368}
        & \textbf{0.5038} \\
        \bottomrule
    \end{tabular}
    \vspace{-2mm}
\end{wraptable}

We evaluate the contribution of CauKerV2 by removing its
\(10\%\) sampling share and redistributing it to
GIFT-Eval-Pretrain+. Its shares therefore increase
from \(25\%\) to \(35\%\) each, while KernelSynth remains at
\(30\%\). All other settings remain unchanged, including
the 110K-step training budget.

As shown in Table~\ref{tab:caukerv2-ablation}, removing CauKerV2
increases MASE and CRPS by \(2.29\%\) and \(1.42\%\),
respectively. Replacing CauKerV2 with additional real-world
samples therefore degrades both point and probabilistic
forecasting performance. These results suggest that CauKerV2
provides a complementary training signal beyond simply
increasing the proportion of real-world data.

\subsection{Analysis of Terminal Mask Length}
\label{sec:mask_ablation}

We examine the effect of longer extrapolation targets during
pretraining by increasing the maximum terminal mask length
from 2 to 32 patches. Both variants use \smallmodelname{}
and the same 110K-step training budget, with all other
settings unchanged.

\begin{wraptable}{r}{0.35\textwidth}
    \vspace{-0.4cm}
    \centering
    \small
    \setlength{\tabcolsep}{3pt}
    \renewcommand{\arraystretch}{1.1}
    \caption{Terminal mask ablation.}
    \label{tab:terminal-mask-ablation}
    \begin{tabular}{@{}lcc@{}}
        \toprule
        Max.\ patches & MASE & CRPS \\
        \midrule
        32
        & 0.7473
        & 0.5116 \\
        2
        & \textbf{0.7368}
        & \textbf{0.5038} \\
        \bottomrule
    \end{tabular}
\end{wraptable}
As shown in Table~\ref{tab:terminal-mask-ablation}, increasing
the terminal mask length raises MASE and CRPS by \(1.43\%\)
and \(1.55\%\), respectively. Thus, longer extrapolation
targets do not improve forecasting performance under the
same training budget. These results support restricting
terminal masks to short spans within our hybrid masking
strategy.

\section{Limitations}
\label{sec:limitations}

Our study has several limitations. First, \modelname{} adopts a channel-independent architecture and does not explicitly model cross-variable dependencies or external covariates.
Second, computational constraints limit the scope of our ablations. Controlled comparisons are conducted primarily with \smallmodelname{} at a fixed training budget. 
Third, our data mixture and synthetic generators encode manually chosen priors. Although the CauKerV2 ablation supports its complementarity to real world data, it does not isolate the
contributions of individual generator families, structural composition, or online sampling.

\section{Conclusion}
\label{sec:conclusion}

We presented \modelname, a time series foundation model built on patch Transformer, together with an open pretraining recipe covering data construction, training, and evaluation.
The recipe combines curated real world data with KernelSynth and CauKerV2, a multi-stage learning-rate schedule that produces reusable intermediate checkpoints, and quantile deep supervision
that directly trains intermediate predictions.

Experiments demonstrate competitive forecasting, classification transfer through
frozen representations, and zero-shot anomaly detection. Controlled ablations support the
contributions of progressive convergence learning rate schedule, deep quantile supervision, and CauKerV2. By releasing the model and the complete pipeline, we aim to make TSFM pretraining easier
to reproduce and extend, and to provide a foundation for further research on synthetic data, training objectives.

\bibliography{main}
\bibliographystyle{tmlr}

\appendix
\newpage
\section{Model Architecture Details}
\label{app:model-architecture-details}

\subsection{Input window and mask-aware normalization}

Let
\[
n_o=\sum_{t=1}^{L_{\max}}o_t,
\qquad
\bar n_o=\max(n_o,1),
\]
and compute
\[
\mu
=
\frac{1}{\bar n_o}
\sum_{t=1}^{L_{\max}}o_t x_t,
\qquad
v
=
\frac{1}{\bar n_o}
\sum_{t=1}^{L_{\max}}o_t(x_t-\mu)^2.
\]
The scale used by the released training and evaluation implementation is
\[
\bar s
=
\begin{cases}
\sqrt{v+\epsilon}, & n_o\ge 2,\\
1,                  & n_o<2,
\end{cases}
\qquad
\epsilon=10^{-5}.
\]
The normalized input and inverse transformation are
\[
z_t
=
\operatorname{asinh}
\left(
\frac{x_t-\mu}{\bar s}
\right),
\qquad
\widetilde z_t=o_tz_t,
\]
\[
\hat x_{t,\tau}
=
\mu+\bar s\,\sinh(\hat z_{t,\tau}).
\]

% Let $L$ denote the complete model window, including both observed and masked
% positions. For one channel, let
% $\mathbf{x}=(x_1,\ldots,x_L)\in\mathbb{R}^{L}$ and define
% \begin{equation}
%     m_t
%     =m_t^{\mathrm{pred}}\lor m_t^{\mathrm{miss}}\lor m_t^{\mathrm{pad}},
%     \qquad
%     o_t=1-m_t,
% \end{equation}
% where $m_t=1$ means that position $t$ is withheld, missing, or padded, and
% $o_t$ is its visibility indicator. Normalization statistics are computed only
% from visible observations:
% \begin{align}
%     n_o &= \max\!\left(\sum_{t=1}^{L}o_t,1\right), \\
%     \mu &= \frac{1}{n_o}\sum_{t=1}^{L}o_t x_t, \\
%     s &= \sqrt{\frac{1}{n_o}
%         \sum_{t=1}^{L}o_t(x_t-\mu)^2}, \\
%     \bar{s} &=
%     \begin{cases}
%         s, & s>\epsilon,\\
%         1, & s\leq\epsilon,
%     \end{cases}
%     \qquad \epsilon=10^{-5}.
% \end{align}
% The model uses an inverse-hyperbolic-sine transform after standardization:
% \begin{equation}
%     z_t=\operatorname{asinh}\!\left(\frac{x_t-\mu}{\bar{s}}\right),
%     \qquad
%     \widetilde{z}_t=o_t z_t.
% \end{equation}
% Thus, masked positions are set to zero in normalized space and are explicitly
% distinguished from genuine zeros by $o_t$. A normalized prediction
% $\widehat{z}_{t,k}$ is mapped back to the original scale as
% \begin{equation}
%     \widehat{x}_{t,k}
%     =\mu+\bar{s}\,\sinh(\widehat{z}_{t,k}).
% \end{equation}

\subsection{Patch encoder}

The window is split into $N=L/P$ non-overlapping patches of length $P$. For
patch $n$, we concatenate its normalized values and visibility indicators:
\begin{equation}
    \mathbf{u}_n
    =\left[
        \widetilde{\mathbf{z}}_n;
        \mathbf{o}_n
      \right]
    \in\mathbb{R}^{2P}.
\end{equation}
Each patch is mapped to the model dimension by a residual two-layer projection,
\begin{equation}
    \phi_{\mathrm{in}}(\mathbf{u})
    =\mathbf{W}_2\,
      \sigma(\mathbf{W}_1\mathbf{u}+\mathbf{b}_1)
      +\mathbf{b}_2
      +\mathbf{W}_{r}\mathbf{u}+\mathbf{b}_{r},
\end{equation}
where $\sigma$ is the sigmoid function. The initial representation is
\begin{equation}
    \mathbf{h}^{(0)}_n
    =\phi_{\mathrm{in}}(\mathbf{u}_n)+\mathbf{e}_n,
\end{equation}
where $\mathbf{e}_n\in\mathbb{R}^{d}$ is a learned absolute positional
embedding. Padded patches are excluded as attention keys, while attention among
all non-padded patches is bidirectional and non-causal.

\subsection{Transformer encoder}

The encoder contains $M$ pre-normalization Transformer blocks. For layer
$\ell=1,\ldots,M$,
\begin{align}
    \mathbf{a}^{(\ell)}
    &=\mathbf{h}^{(\ell-1)}
      +\operatorname{MHA}\!\left(
        \operatorname{LN}(\mathbf{h}^{(\ell-1)})
       \right), \\
    \mathbf{h}^{(\ell)}
    &=\mathbf{a}^{(\ell)}
      +\operatorname{Dropout}\!\left(
        \operatorname{FFN}\!\left(
          \operatorname{LN}(\mathbf{a}^{(\ell)})
        \right)
       \right).
\end{align}
Multi-head self-attention uses $n_h=d/d_h$ heads. The feed-forward network has
hidden width $4d$ and a GELU activation. No causal mask is used; information
leakage is prevented by zeroing the withheld values and providing only their
visibility indicators to the encoder.

\subsection{Monotone quantile decoder}

Let $K$ be the number of predicted quantiles. A residual output projection maps
each final patch representation to $P(K+1)$ raw values. After reshaping, let
$r_{n,j,0},\ldots,r_{n,j,K}$ denote the raw outputs for offset $j$ within patch
$n$. The normalized quantile predictions are
\begin{equation}
    \widehat{z}_{n,j,k}
    =r_{n,j,0}
     +\sum_{i=1}^{k}
       \frac{\operatorname{softplus}(r_{n,j,i})}{K},
    \qquad k=1,\ldots,K.
\end{equation}
Because every increment is positive,
$\widehat{z}_{n,j,k+1}\geq\widehat{z}_{n,j,k}$, so quantile crossing cannot
occur. With $K=99$, the modeled levels are
\begin{equation}
    \tau_k=\frac{k}{K+1}=\frac{k}{100},
    \qquad k=1,\ldots,99,
\end{equation}
covering $0.01$ through $0.99$. These outputs represent pointwise conditional
marginal quantiles; the model does not define a joint distribution over the
forecast horizon.

\subsection{Forecast-window construction and multivariate inputs}

For a requested forecast horizon $H$, the inference wrapper reserves
\begin{equation}
    F=\max\!\left(H,\,P\max(B_{\mathrm{cpm}},2)\right)
     =\max(H,128)
\end{equation}
masked positions, where $P=16$ and $B_{\mathrm{cpm}}=8$. It retains at most
$L-F$ recent observations, left-pads shorter inputs, appends the $F$ masked
positions, and returns the first $H$ forecasts. Consequently, $L=8192$ is the
total model window rather than the number of historical observations always
available to the encoder.

For a multivariate input
$\mathbf{X}\in\mathbb{R}^{B\times L\times D_v}$, the implementation reshapes
the input to $(BD_v)\times L$, applies the same univariate backbone to every
channel, and then restores the channel dimension. Normalization statistics are
computed separately for each channel. Hence, all channels share model
parameters, but there is no cross-channel attention in the current
architecture.

% When a point forecast is required, the implementation aggregates the predicted
% quantiles using symmetric triangular weights:
% \begin{equation}
%     \widehat{x}^{\mathrm{point}}_t
%     =\sum_{k=1}^{K}w_k\widehat{x}_{t,k},
%     \qquad
%     w_k=
%     \frac{0.5-|0.5-\tau_k|}
%          {\sum_{i=1}^{K}\left(0.5-|0.5-\tau_i|\right)}.
% \end{equation}

The inference interface optionally provides a single-value summary of the
predicted quantiles using symmetric triangular weights:
\begin{equation}
    \widehat{y}^{\mathrm{tri}}_{t}
    =
    \sum_{k=1}^{K}
    w_k \widehat{q}_{t,\tau_k},
    \qquad
    w_k
    =
    \frac{0.5-\lvert 0.5-\tau_k\rvert}
    {\sum_{i=1}^{K}
    \left(0.5-\lvert 0.5-\tau_i\rvert\right)}.
    \label{eq:triangular-point-summary}
\end{equation}
This aggregation assigns larger weights to quantiles near the center of the
predictive distribution. It is provided as an optional point summary and
should not be interpreted as the predictive mean. In all forecasting results
reported in this work, including MASE, we instead use the median forecast,
\begin{equation}
    \widehat{y}^{\mathrm{median}}_{t}
    =
    \widehat{q}_{t,0.5}.
    \label{eq:median-point-forecast}
\end{equation}

\subsection{Configuration and parameter count}

\begin{table}[t]
    \centering
    \small
    \caption{Architecture configuration of \modelname.}
    \label{tab:tabby-architecture-config}
    \begin{tabular}{@{}lc@{}}
        \toprule
        Hyperparameter & Value \\
        \midrule
        Complete model window $L$ & $8192$ \\
        Patch length $P$ & $16$ \\
        Number of patch tokens $N=L/P$ & $512$ \\
        Transformer layers $M$ & $20$ \\
        Model dimension $d$ & $768$ \\
        Attention heads $n_h$ & $12$ \\
        Head dimension $d_h$ & $64$ \\
        Feed-forward width & $3072$ \\
        Dropout & $0.1$ \\
        Quantile levels $K$ & $99$ \\
        Pretraining mask ratio & $0.4$ \\
        Contiguous-mask setting $B_{\mathrm{cpm}}$ & $8$ patches \\
        Normalization stabilizer $\epsilon$ & $10^{-5}$ \\
        \bottomrule
    \end{tabular}
\end{table}

The exact parameter decomposition is reported in
Table~\ref{tab:tabby-parameter-count}. The input projection receives $2P=32$
features per patch, and the output projection produces
$P(K+1)=1600$ raw values per patch.

\begin{table}[t]
    \centering
    \small
    \caption{Parameter decomposition of \modelname.}
    \label{tab:tabby-parameter-count}
    \begin{tabular}{@{}lrr@{}}
        \toprule
        Component & Parameters & Share \\
        \midrule
        Input residual projection & $641{,}280$ & $0.44\%$ \\
        Learned positional embedding & $393{,}216$ & $0.27\%$ \\
        $20$ Transformer blocks & $141{,}757{,}440$ & $97.20\%$ \\
        Output residual projection & $3{,}051{,}392$ & $2.09\%$ \\
        \midrule
        Total & $145{,}843{,}328$ & $100.00\%$ \\
        \bottomrule
    \end{tabular}
\end{table}

The state dictionary contains $253$ parameter tensors. Storing the parameters
alone requires approximately $556.348$ MiB in FP32 or $278.174$ MiB in BF16;
these figures exclude gradients, optimizer states, and activations.

% Required packages: \usepackage{booktabs,longtable,pdflscape,array}
% Insert this file after \appendix with: \input{appendix_gift_eval_pretrain_plus}

\section{GIFT-Eval-Pretrain+: Real-World Data Corpora}
\label{app:gift-eval-pretrain-plus}

\paragraph{Corpus scope.}
GIFT-Eval-Pretrain+ is the curated real-world component used in our pretraining corpora. It starts from the stored GIFT-Eval-Pretrain collection~\citep{aksu2024gift}, restores a small set of eligible real-world series removed during earlier filtering, and adds complementary datasets selected from the TiRex-2 training corpora~\citep{tirex2}. The final on-disk corpus contains 150 source-dataset entries organized into 134 training sampling units. The difference arises because several small, semantically related datasets are stored in a shared directory and distinguished by a \texttt{source\_dataset} field. These consolidated directories are sampled as single training units under the balanced sampling scheme described in the main text.

\paragraph{Counting convention.}
For dataset $d$, let $N_d$ denote the number of stored Hugging Face rows (samples). If sample $n$ contains $C_{dn}$ channels of length $T_{dn}$, we report
\[
C_d=\sum_{n=1}^{N_d} C_{dn},
\qquad
P_d=\sum_{n=1}^{N_d} C_{dn}T_{dn},
\qquad
\overline{T}_d=\frac{P_d}{C_d}.
\]
Here, $C_d$ is the number of univariate channel sequences, $P_d$ is the number of scalar observations, and $\overline{T}_d$ is the channel-weighted mean sequence length. This distinction is necessary because one stored sample may contain multiple channels. In Table~\ref{tab:gift-plus-inventory}, U denotes entries with $C_d=N_d$, while M denotes entries with $C_d>N_d$. The U/M flag therefore describes the stored target representation rather than the forecasting protocol used downstream.

\paragraph{Overall characteristics.}
The corpus contains 3,792,438 samples, 28,872,274 channel sequences, and 229,802,174,843 scalar observations. The global channel-weighted mean length is 7,959.27 observations. Dataset-level mean lengths range from 48.95 to 538,725.31 observations, covering both short forecasting collections and very long sensor or power-generation records. Seventy-three stored dataset entries contain multichannel samples; together, they contribute 203,407,137,243 observations (88.5\% of the corpus). The collection spans energy, climate and environment, transportation, web and cloud operations, sales, economics and finance, healthcare, tourism, and natural phenomena, with sampling intervals ranging from seconds to quarterly observations. The scan found no malformed targets.

All counts below describe the post-curation files actually used by our data loader; they need not match the raw upstream releases because filtering, restoration, sharding, and channel expansion are applied before pretraining.

\subsection{Consolidation into Training Sampling Units}
\label{app:gift-plus-consolidation}

Uniform sampling over source datasets can severely oversample very small collections. We therefore consolidate selected small datasets before constructing the balanced sampler. The consolidation is performed only at the sampling-unit level: the original dataset identity is retained in the \texttt{source\_dataset} field, and all samples remain individually recoverable. Table~\ref{tab:gift-plus-merged-units} reports every composite unit found in the final training manifest.

The first three units group datasets from the same or closely related application areas. The recovered-real unit instead groups datasets by curation provenance: these datasets were restored after an earlier filtering stage and are not assumed to belong to a single application domain.

\begin{longtable}{>{\raggedright\arraybackslash}p{4.5cm} >{\raggedright\arraybackslash}p{8.1cm} r}
\caption{Source datasets consolidated into shared training sampling units. The row count is the number of stored Hugging Face samples in the composite unit.}
\label{tab:gift-plus-merged-units}\\
\toprule
Training sampling unit & Constituent source datasets & Rows \\
\midrule
\endfirsthead
\toprule
Training sampling unit & Constituent source datasets & Rows \\
\midrule
\endhead
\bottomrule
\endfoot
\texttt{energy\_covariates\_small} & \texttt{cockatoo}, \texttt{covid19\_energy}, \texttt{gfc14\_load}, \texttt{pdb}, \texttt{spain}, and \texttt{spanish\_energy\_and\_weather} & 129 \\
\texttt{macro\_finance\_small} & \texttt{bitcoin\_with\_missing}, \texttt{exchange\_rate}, and \texttt{monash\_fred\_md} & 321 \\
\texttt{public\_health\_small} & \texttt{cdc\_fluview\_ilinet}, \texttt{cdc\_fluview\_who\_nrevss}, and \texttt{covid\_mobility} & 1,304 \\
\texttt{recovered\_useful\_real} & \texttt{HZMETRO}, \texttt{SHMETRO}, \texttt{borealis}, \texttt{elecdemand}, \texttt{elf}, \texttt{smart}, \texttt{solar\_power}, and \texttt{wind\_power} & 4,096 \\
\end{longtable}

\begin{longtable}{clrrrr}
\caption{Composition of GIFT-Eval-Pretrain+ by collection source. Mean length is computed as total observations divided by total channel sequences.}
\label{tab:gift-plus-source-summary}\\
\toprule
Code & Collection component & Entries & Samples & Channels & Observations \\
\midrule
G & Stored GIFT-Eval subset & 120 & 3,782,761 & 28,855,468 & 228,829,433,709 \\
T & TiRex-2 supplement & 21 & 5,474 & 12,603 & 943,028,306 \\
R & Restored curated subset & 9 & 4,203 & 4,203 & 29,712,828 \\
\midrule
Total & GIFT-Eval-Pretrain+ & 150 & 3,792,438 & 28,872,274 & 229,802,174,843 \\ 
\bottomrule
\end{longtable}

\paragraph{Frequency notation.}
\textit{s}, \textit{min}, \textit{h}, \textit{d}, \textit{wk}, \textit{mo}, and \textit{qtr} denote seconds, minutes, hours, days, weeks, months, and quarters, respectively. Frequencies record the nominal sampling interval of the stored source dataset; they are not resampling factors applied by the model.

\begin{landscape}
\scriptsize
\setlength{\tabcolsep}{3pt}
\renewcommand{\arraystretch}{1.05}
\begin{longtable}{>{\raggedright\arraybackslash}p{5.4cm} c >{\raggedright\arraybackslash}p{3.0cm} c c r r r r}
\caption{Complete dataset-level inventory of GIFT-Eval-Pretrain+. Source codes follow Table~\ref{tab:gift-plus-source-summary}. Samples are stored rows; channels are univariate sequences after expanding the target dimension; mean length is $P_d/C_d$; and observations count every scalar value across all channels.}
\label{tab:gift-plus-inventory}\\
\toprule
Dataset & Src. & Domain & Freq. & Type & Samples & Channels & Mean length & Observations \\
\midrule
\endfirsthead
\multicolumn{9}{c}{\tablename\ \thetable{} -- continued from previous page}\\
\toprule
Dataset & Src. & Domain & Freq. & Type & Samples & Channels & Mean length & Observations \\
\midrule
\endhead
\midrule
\multicolumn{9}{r}{Continued on next page}\\
\endfoot
\bottomrule
\endlastfoot
\texttt{BEIJING\_SUBWAY\_30MIN} & T & Transport & 30 min & M & 276 & 552 & 1,572.00 & 867,744 \\
\texttt{PEMS04} & T & Transport & 5 min & M & 307 & 921 & 16,992.00 & 15,649,632 \\
\texttt{PEMS08} & T & Transport & 5 min & M & 170 & 510 & 17,856.00 & 9,106,560 \\
\texttt{beijing\_air\_quality} & T & Environment & 1 h & M & 12 & 132 & 35,064.00 & 4,628,448 \\
\texttt{china\_air\_quality} & T & Environment & 1 h & M & 437 & 2,622 & 13,133.26 & 34,435,404 \\
\texttt{cockatoo} & T & Energy & 1 h & U & 6 & 6 & 17,544.00 & 105,264 \\
\texttt{covid19\_energy} & T & Energy & 1 h & U & 14 & 14 & 31,912.00 & 446,768 \\
\texttt{gfc14\_load} & T & Energy & 1 h & U & 1 & 1 & 17,520.00 & 17,520 \\
\texttt{pdb} & T & Energy & 1 h & U & 2 & 2 & 17,520.00 & 35,040 \\
\texttt{spain} & T & Energy & 1 h & U & 4 & 4 & 35,064.00 & 140,256 \\
\texttt{spanish\_energy\_and\_weather} & T & Energy & 1 h & U & 102 & 102 & 35,064.00 & 3,576,528 \\
\texttt{bitcoin\_with\_missing} & R & Econ./finance & 1 d & U & 107 & 107 & 4,551.00 & 486,957 \\
\texttt{exchange\_rate} & T & Econ./finance & 1 d & U & 107 & 107 & 7,588.00 & 811,916 \\
\texttt{monash\_fred\_md} & T & Econ./finance & 1 mo & U & 107 & 107 & 728.00 & 77,896 \\
\texttt{project\_tycho} & T & Healthcare & 1 wk & U & 1,258 & 1,258 & 1,095.16 & 1,377,707 \\
\texttt{cdc\_fluview\_ilinet} & T & Healthcare & 1 wk & U & 646 & 646 & 887.00 & 573,000 \\
\texttt{cdc\_fluview\_who\_nrevss} & T & Healthcare & 1 wk & U & 296 & 296 & 564.32 & 167,040 \\
\texttt{covid\_mobility} & T & Transport & 1 d & U & 362 & 362 & 410.50 & 148,602 \\
\texttt{HZMETRO} & R & Transport & 15 min & U & 512 & 512 & 2,377.00 & 1,217,024 \\
\texttt{SHMETRO} & R & Transport & 15 min & U & 512 & 512 & 8,192.00 & 4,194,304 \\
\texttt{borealis} & R & Energy & 1 h & U & 512 & 512 & 5,552.78 & 2,843,023 \\
\texttt{elecdemand} & R & Energy & 30 min & U & 512 & 512 & 8,192.00 & 4,194,304 \\
\texttt{elf} & R & Energy & 1 h & U & 512 & 512 & 8,192.00 & 4,194,304 \\
\texttt{smart} & R & Energy & 1 h & U & 512 & 512 & 8,192.00 & 4,194,304 \\
\texttt{solar\_power} & R & Energy & 4 s & U & 512 & 512 & 8,192.00 & 4,194,304 \\
\texttt{wind\_power} & R & Energy & 4 s & U & 512 & 512 & 8,192.00 & 4,194,304 \\
\texttt{residential\_load\_power} & T & Energy & 1 min & M & 271 & 813 & 538,725.31 & 437,983,677 \\
\texttt{residential\_pv\_power} & T & Energy & 1 min & M & 233 & 699 & 537,935.41 & 376,016,850 \\
\texttt{subseasonal} & T & Climate & 1 d & M & 862 & 3,448 & 16,470.00 & 56,788,560 \\
\texttt{sunspot\_with\_missing} & T & Nature & 1 d & U & 1 & 1 & 73,894.00 & 73,894 \\
\texttt{mexico\_city\_bikes} & G & Transport & 1 h & U & 494 & 494 & 78,313.77 & 38,687,004 \\
\texttt{ushcn\_daily} & G & Climate & 1 d & U & 1,218 & 1,218 & 38,653.62 & 47,080,115 \\
\texttt{weatherbench\_daily} & G & Climate & 1 d & U & 225,280 & 225,280 & 14,609.98 & 3,291,336,704 \\
\texttt{weatherbench\_weekly} & G & Climate & 1 wk & U & 225,280 & 225,280 & 2,087.00 & 470,159,360 \\
\texttt{LOS\_LOOP} & G & Transport & 5 min & U & 207 & 207 & 34,272.00 & 7,094,304 \\
\texttt{PEMS03} & G & Transport & 5 min & U & 358 & 358 & 26,208.00 & 9,382,464 \\
\texttt{PEMS07} & G & Transport & 5 min & U & 883 & 883 & 28,224.00 & 24,921,792 \\
\texttt{PEMS\_BAY} & G & Transport & 5 min & U & 325 & 325 & 52,128.00 & 16,941,600 \\
\texttt{Q-TRAFFIC} & G & Transport & 15 min & U & 45,148 & 45,148 & 5,856.00 & 264,386,688 \\
\texttt{alibaba\_cluster\_trace\_2018} & G & Cloud operations & 5 min & M & 58,409 & 116,818 & 1,629.76 & 190,385,060 \\
\texttt{australian\_electricity\_demand} & G & Energy & 30 min & U & 5 & 5 & 230,716.80 & 1,153,584 \\
\texttt{azure\_vm\_traces\_2017} & G & Cloud operations & 5 min & U & 159,472 & 159,472 & 5,552.84 & 885,522,908 \\
\texttt{bdg-2\_bear} & G & Energy & 1 h & U & 91 & 91 & 16,289.14 & 1,482,312 \\
\texttt{bdg-2\_fox} & G & Energy & 1 h & U & 135 & 135 & 17,219.02 & 2,324,568 \\
\texttt{bdg-2\_panther} & G & Energy & 1 h & U & 105 & 105 & 8,760.00 & 919,800 \\
\texttt{bdg-2\_rat} & G & Energy & 1 h & U & 280 & 280 & 16,886.74 & 4,728,288 \\
\texttt{borg\_cluster\_data\_2011} & G & Cloud operations & 5 min & M & 143,386 & 286,772 & 3,748.99 & 1,075,105,708 \\
\texttt{buildings\_900k} & G & Energy & 1 h & U & 1,795,256 & 1,795,256 & 8,761.00 & 15,728,237,816 \\
\texttt{bull} & G & Energy & 1 h & U & 41 & 41 & 17,544.00 & 719,304 \\
\texttt{cmip6\_1850} & G & Climate & 6 h & M & 8,192 & 434,176 & 7,300.00 & 3,169,484,800 \\
\texttt{cmip6\_1855} & G & Climate & 6 h & M & 8,192 & 434,176 & 7,300.00 & 3,169,484,800 \\
\texttt{cmip6\_1860} & G & Climate & 6 h & M & 8,192 & 434,176 & 7,300.00 & 3,169,484,800 \\
\texttt{cmip6\_1865} & G & Climate & 6 h & M & 8,192 & 434,176 & 7,300.00 & 3,169,484,800 \\
\texttt{cmip6\_1870} & G & Climate & 6 h & M & 8,192 & 434,176 & 7,300.00 & 3,169,484,800 \\
\texttt{cmip6\_1875} & G & Climate & 6 h & M & 8,192 & 434,176 & 7,300.00 & 3,169,484,800 \\
\texttt{cmip6\_1880} & G & Climate & 6 h & M & 8,192 & 434,176 & 7,300.00 & 3,169,484,800 \\
\texttt{cmip6\_1885} & G & Climate & 6 h & M & 8,192 & 434,176 & 7,300.00 & 3,169,484,800 \\
\texttt{cmip6\_1890} & G & Climate & 6 h & M & 8,192 & 434,176 & 7,300.00 & 3,169,484,800 \\
\texttt{cmip6\_1895} & G & Climate & 6 h & M & 8,192 & 434,176 & 7,300.00 & 3,169,484,800 \\
\texttt{cmip6\_1900} & G & Climate & 6 h & M & 8,192 & 434,176 & 7,300.00 & 3,169,484,800 \\
\texttt{cmip6\_1905} & G & Climate & 6 h & M & 8,192 & 434,176 & 7,300.00 & 3,169,484,800 \\
\texttt{cmip6\_1910} & G & Climate & 6 h & M & 8,192 & 434,176 & 7,300.00 & 3,169,484,800 \\
\texttt{cmip6\_1915} & G & Climate & 6 h & M & 8,192 & 434,176 & 7,300.00 & 3,169,484,800 \\
\texttt{cmip6\_1920} & G & Climate & 6 h & M & 8,192 & 434,176 & 7,300.00 & 3,169,484,800 \\
\texttt{cmip6\_1925} & G & Climate & 6 h & M & 8,192 & 434,176 & 7,300.00 & 3,169,484,800 \\
\texttt{cmip6\_1930} & G & Climate & 6 h & M & 8,192 & 434,176 & 7,300.00 & 3,169,484,800 \\
\texttt{cmip6\_1935} & G & Climate & 6 h & M & 8,192 & 434,176 & 7,300.00 & 3,169,484,800 \\
\texttt{cmip6\_1940} & G & Climate & 6 h & M & 8,192 & 434,176 & 7,300.00 & 3,169,484,800 \\
\texttt{cmip6\_1945} & G & Climate & 6 h & M & 8,192 & 434,176 & 7,300.00 & 3,169,484,800 \\
\texttt{cmip6\_1950} & G & Climate & 6 h & M & 8,192 & 434,176 & 7,300.00 & 3,169,484,800 \\
\texttt{cmip6\_1955} & G & Climate & 6 h & M & 8,192 & 434,176 & 7,300.00 & 3,169,484,800 \\
\texttt{cmip6\_1960} & G & Climate & 6 h & M & 8,192 & 434,176 & 7,300.00 & 3,169,484,800 \\
\texttt{cmip6\_1965} & G & Climate & 6 h & M & 8,192 & 434,176 & 7,300.00 & 3,169,484,800 \\
\texttt{cmip6\_1970} & G & Climate & 6 h & M & 8,192 & 434,176 & 7,300.00 & 3,169,484,800 \\
\texttt{cmip6\_1975} & G & Climate & 6 h & M & 8,192 & 434,176 & 7,300.00 & 3,169,484,800 \\
\texttt{cmip6\_1980} & G & Climate & 6 h & M & 8,192 & 434,176 & 7,300.00 & 3,169,484,800 \\
\texttt{cmip6\_1985} & G & Climate & 6 h & M & 8,192 & 434,176 & 7,300.00 & 3,169,484,800 \\
\texttt{cmip6\_1990} & G & Climate & 6 h & M & 8,192 & 434,176 & 7,300.00 & 3,169,484,800 \\
\texttt{cmip6\_1995} & G & Climate & 6 h & M & 8,192 & 434,176 & 7,300.00 & 3,169,484,800 \\
\texttt{cmip6\_2000} & G & Climate & 6 h & M & 8,192 & 434,176 & 7,300.00 & 3,169,484,800 \\
\texttt{cmip6\_2005} & G & Climate & 6 h & M & 8,192 & 434,176 & 7,300.00 & 3,169,484,800 \\
\texttt{cmip6\_2010} & G & Climate & 6 h & M & 8,192 & 434,176 & 7,300.00 & 3,169,484,800 \\
\texttt{era5\_1989} & G & Climate & 1 h & M & 8,192 & 368,640 & 8,736.00 & 3,220,439,040 \\
\texttt{era5\_1990} & G & Climate & 1 h & M & 8,192 & 368,640 & 8,736.00 & 3,220,439,040 \\
\texttt{era5\_1991} & G & Climate & 1 h & M & 8,192 & 368,640 & 8,736.00 & 3,220,439,040 \\
\texttt{era5\_1992} & G & Climate & 1 h & M & 8,192 & 368,640 & 8,736.00 & 3,220,439,040 \\
\texttt{era5\_1993} & G & Climate & 1 h & M & 8,192 & 368,640 & 8,736.00 & 3,220,439,040 \\
\texttt{era5\_1994} & G & Climate & 1 h & M & 8,192 & 368,640 & 8,736.00 & 3,220,439,040 \\
\texttt{era5\_1995} & G & Climate & 1 h & M & 8,192 & 368,640 & 8,736.00 & 3,220,439,040 \\
\texttt{era5\_1996} & G & Climate & 1 h & M & 8,192 & 368,640 & 8,736.00 & 3,220,439,040 \\
\texttt{era5\_1997} & G & Climate & 1 h & M & 8,192 & 368,640 & 8,736.00 & 3,220,439,040 \\
\texttt{era5\_1998} & G & Climate & 1 h & M & 8,192 & 368,640 & 8,736.00 & 3,220,439,040 \\
\texttt{era5\_1999} & G & Climate & 1 h & M & 8,192 & 368,640 & 8,736.00 & 3,220,439,040 \\
\texttt{era5\_2000} & G & Climate & 1 h & M & 8,192 & 368,640 & 8,736.00 & 3,220,439,040 \\
\texttt{era5\_2001} & G & Climate & 1 h & M & 8,192 & 368,640 & 8,736.00 & 3,220,439,040 \\
\texttt{era5\_2002} & G & Climate & 1 h & M & 8,192 & 368,640 & 8,736.00 & 3,220,439,040 \\
\texttt{era5\_2003} & G & Climate & 1 h & M & 8,192 & 368,640 & 8,736.00 & 3,220,439,040 \\
\texttt{era5\_2004} & G & Climate & 1 h & M & 8,192 & 368,640 & 8,736.00 & 3,220,439,040 \\
\texttt{era5\_2005} & G & Climate & 1 h & M & 8,192 & 368,640 & 8,736.00 & 3,220,439,040 \\
\texttt{era5\_2006} & G & Climate & 1 h & M & 8,192 & 368,640 & 8,736.00 & 3,220,439,040 \\
\texttt{era5\_2007} & G & Climate & 1 h & M & 8,192 & 368,640 & 8,736.00 & 3,220,439,040 \\
\texttt{era5\_2008} & G & Climate & 1 h & M & 8,192 & 368,640 & 8,736.00 & 3,220,439,040 \\
\texttt{era5\_2009} & G & Climate & 1 h & M & 8,192 & 368,640 & 8,736.00 & 3,220,439,040 \\
\texttt{era5\_2010} & G & Climate & 1 h & M & 8,192 & 368,640 & 8,736.00 & 3,220,439,040 \\
\texttt{era5\_2011} & G & Climate & 1 h & M & 8,192 & 368,640 & 8,736.00 & 3,220,439,040 \\
\texttt{era5\_2012} & G & Climate & 1 h & M & 8,192 & 368,640 & 8,736.00 & 3,220,439,040 \\
\texttt{era5\_2013} & G & Climate & 1 h & M & 8,192 & 368,640 & 8,736.00 & 3,220,439,040 \\
\texttt{era5\_2014} & G & Climate & 1 h & M & 8,192 & 368,640 & 8,736.00 & 3,220,439,040 \\
\texttt{era5\_2015} & G & Climate & 1 h & M & 8,192 & 368,640 & 8,736.00 & 3,220,439,040 \\
\texttt{era5\_2016} & G & Climate & 1 h & M & 8,192 & 368,640 & 8,736.00 & 3,220,439,040 \\
\texttt{era5\_2017} & G & Climate & 1 h & M & 8,192 & 368,640 & 8,736.00 & 3,220,439,040 \\
\texttt{era5\_2018} & G & Climate & 1 h & M & 8,192 & 368,640 & 8,736.00 & 3,220,439,040 \\
\texttt{extended\_web\_traffic\_with\_missing} & G & Web & 1 d & U & 145,063 & 145,063 & 2,557.00 & 370,926,091 \\
\texttt{favorita\_sales} & G & Sales/retail & 1 d & U & 111,840 & 111,840 & 1,244.45 & 139,179,538 \\
\texttt{favorita\_transactions} & G & Sales/retail & 1 d & U & 54 & 54 & 1,563.11 & 84,408 \\
\texttt{gfc12\_load} & G & Energy & 1 h & U & 20 & 20 & 39,414.00 & 788,280 \\
\texttt{gfc17\_load} & G & Energy & 1 h & U & 8 & 8 & 17,544.00 & 140,352 \\
\texttt{hog} & G & Energy & 1 h & U & 24 & 24 & 17,544.00 & 421,056 \\
\texttt{ideal} & G & Energy & 1 h & U & 217 & 217 & 5,784.58 & 1,255,253 \\
\texttt{kaggle\_web\_traffic\_weekly} & G & Web & 1 wk & U & 145,063 & 145,063 & 114.00 & 16,537,182 \\
\texttt{kdd2022} & G & Energy & 10 min & U & 134 & 134 & 35,279.99 & 4,727,519 \\
\texttt{largest\_2017} & G & Transport & 5 min & U & 8,196 & 8,196 & 105,120.00 & 861,563,520 \\
\texttt{largest\_2018} & G & Transport & 5 min & U & 8,428 & 8,428 & 105,120.00 & 885,951,360 \\
\texttt{largest\_2019} & G & Transport & 5 min & U & 8,600 & 8,600 & 105,120.00 & 904,032,000 \\
\texttt{largest\_2020} & G & Transport & 5 min & U & 8,561 & 8,561 & 105,408.00 & 902,397,888 \\
\texttt{largest\_2021} & G & Transport & 5 min & U & 8,548 & 8,548 & 105,120.00 & 898,565,760 \\
\texttt{lcl} & G & Energy & 1 h & U & 713 & 713 & 13,385.07 & 9,543,553 \\
\texttt{london\_smart\_meters\_with\_missing} & G & Energy & 30 min & U & 5,520 & 5,520 & 30,115.74 & 166,238,880 \\
\texttt{m1\_monthly} & G & Econ./finance & 1 mo & U & 617 & 617 & 72.76 & 44,892 \\
\texttt{m5} & G & Sales/retail & 1 d & U & 30,490 & 30,490 & 1,913.00 & 58,327,370 \\
\texttt{monash\_m3\_monthly} & G & Econ./finance & 1 mo & U & 1,428 & 1,428 & 99.34 & 141,858 \\
\texttt{monash\_m3\_other} & G & Econ./finance & 1 qtr & U & 174 & 174 & 68.58 & 11,933 \\
\texttt{monash\_m3\_quarterly} & G & Econ./finance & 1 qtr & U & 756 & 756 & 48.95 & 37,004 \\
\texttt{nn5\_daily\_with\_missing} & G & Econ./finance & 1 d & U & 111 & 111 & 735.00 & 81,585 \\
\texttt{nn5\_weekly} & G & Econ./finance & 1 wk & U & 111 & 111 & 105.00 & 11,655 \\
\texttt{oikolab\_weather} & G & Climate & 1 h & U & 8 & 8 & 100,057.00 & 800,456 \\
\texttt{pedestrian\_counts} & G & Transport & 1 h & U & 66 & 66 & 47,435.79 & 3,130,762 \\
\texttt{rideshare\_with\_missing} & G & Transport & 1 h & U & 2,304 & 2,304 & 373.00 & 859,392 \\
\texttt{subseasonal\_precip} & G & Climate & 1 d & U & 862 & 862 & 11,323.00 & 9,760,426 \\
\texttt{taxi\_30min} & G & Transport & 30 min & U & 67,984 & 67,984 & 809.00 & 54,999,056 \\
\texttt{tourism\_monthly} & G & Tourism & 1 mo & U & 366 & 366 & 274.58 & 100,496 \\
\texttt{tourism\_quarterly} & G & Tourism & 1 qtr & U & 427 & 427 & 91.63 & 39,128 \\
\texttt{traffic\_hourly} & G & Transport & 1 h & U & 862 & 862 & 17,376.00 & 14,978,112 \\
\texttt{traffic\_weekly} & G & Transport & 1 wk & U & 862 & 862 & 96.00 & 82,752 \\
\texttt{uber\_tlc\_daily} & G & Transport & 1 d & U & 262 & 262 & 179.72 & 47,087 \\
\texttt{uber\_tlc\_hourly} & G & Transport & 1 h & U & 262 & 262 & 4,310.85 & 1,129,444 \\
\texttt{vehicle\_trips\_with\_missing} & G & Transport & 1 d & U & 329 & 329 & 98.82 & 32,512 \\
\texttt{weather} & G & Climate & 1 d & U & 3,010 & 3,010 & 14,266.35 & 42,941,700 \\
\texttt{wiki-rolling\_nips} & G & Web & 1 d & U & 47,675 & 47,675 & 852.00 & 40,619,100 \\
\texttt{wind\_farms\_with\_missing} & G & Energy & 1 min & U & 337 & 337 & 510,876.47 & 172,165,370 \\
\end{longtable}
\end{landscape}

\section{Configuration of \smallmodelname}
\label{app:tabby-small-configuration}

To reduce the computational cost of controlled ablation studies, we construct
\smallmodelname{} by reducing the depth and hidden dimension of the full
model while retaining its long-context input representation, patch-based
encoder, probabilistic quantile decoder, masking objective, and pretraining
data mixture. The resulting model contains exactly 40{,}291{,}456 parameters
(40.291M), measured directly from the checkpoint at step 110{,}000. The
checkpoint contains 157 parameter tensors with a raw weight size of
153.700 MiB.

The model is trained for 110{,}000 optimization steps on eight GPUs. With a
per-device batch size of 64 and no gradient accumulation, its global batch
size is
\[
B_{\mathrm{global}}
=
8 \times 64 \times 1
=
512.
\]
The total number of sampled training windows is therefore
\[
N_{\mathrm{windows}}
=
110{,}000 \times 512
=
56{,}320{,}000.
\]
This value represents the number of sample exposures during optimization
rather than the number of unique time series, because the stored corpora are
repeatedly sampled and CauKerV2 generates synthetic samples online. The
complete configuration is reported in
Table~\ref{tab:tabby-small-configuration}.

\small
\begin{longtable}{p{0.58\textwidth}p{0.27\textwidth}}
\caption{
Complete architecture, data, masking, supervision, and optimization
configuration of \smallmodelname.
}
\label{tab:tabby-small-configuration}
\\
\toprule
\textbf{Configuration} & \textbf{Value} \\
\midrule
\endfirsthead

\multicolumn{2}{c}{
\textit{Table~\ref{tab:tabby-small-configuration} continued}
}
\\
\toprule
\textbf{Configuration} & \textbf{Value} \\
\midrule
\endhead

\midrule
\multicolumn{2}{r}{\textit{Continued on the next page}} \\
\endfoot

\bottomrule
\endlastfoot

\multicolumn{2}{l}{\textit{Model architecture}} \\
Context length
    & 8{,}192 \\
Patch size
    & 16 \\
Maximum number of patch tokens
    & 512 \\
Number of Transformer layers
    & 12 \\
Hidden dimension
    & 512 \\
Attention-head dimension
    & 32 \\
Number of attention heads
    & 16 \\
MLP expansion ratio
    & 4 \\
MLP hidden dimension
    & 2{,}048 \\
Dropout probability
    & 0.10 \\
Number of predicted quantiles
    & 99 \\
Number of parameter tensors
    & 157 \\
Number of parameters
    & 40{,}291{,}456 \\
Number of parameters in millions
    & 40.291M \\
Raw checkpoint weight size
    & 153.700 MiB \\
\midrule

\multicolumn{2}{l}{\textit{Pretraining data mixture}} \\
BLAST sampling ratio
    & 0.35 \\
KernelSynth sampling ratio
    & 0.30 \\
GIFT-Eval-Pretrain+ sampling ratio
    & 0.25 \\
CauKerV2 sampling ratio
    & 0.10 \\
Mixup sampling ratio
    & 0 \\
Minimum general real-series length
    & 64 \\
BLAST sequence-length range
    & $[96,4096]$ \\
KernelSynth sequence-length range
    & $[96,8192]$ \\
GIFT-Eval-Pretrain+ sequence-length range
    & $[2048,8192]$ \\
CauKerV2 sequence-length range
    & $[96,2048]$ \\
Sequence-length sampling strategy
    & Uniform \\
Length-weighting coefficient
    & 0.5 \\
GIFT-Eval-Pretrain+ balancing
    & Dataset/domain balanced \\
KernelSynth shard balancing
    & Enabled \\
\midrule

\multicolumn{2}{l}{\textit{CauKerV2 configuration}} \\
Generation mode
    & Online \\
Number of generated features
    & 8 \\
Number of SCM nodes
    & 8 \\
Maximum number of parents per node
    & 2 \\
Maximum generated sequence length
    & 2{,}048 \\
\midrule

\multicolumn{2}{l}{\textit{Data-loading configuration}} \\
Number of BLAST workers
    & 3 \\
Number of KernelSynth workers
    & 3 \\
Number of GIFT-Eval-Pretrain+ workers
    & 3 \\
Number of mixup workers
    & 0 \\
Prefetch factor
    & 4 \\
\midrule

\multicolumn{2}{l}{\textit{Masking configuration}} \\
Overall mask ratio
    & 0.40 \\
Number of CPM blocks
    & 8 \\
Minimum terminal-mask length
    & 0 patches \\
Maximum terminal-mask length
    & 3 patches \\
\midrule

\multicolumn{2}{l}{\textit{Deep-supervision configuration}} \\
Deep supervision
    & Enabled \\
Number of supervised exits
    & 5 \\
Decoded exit layers
    & $\{0,3,6,9,12\}$ \\
Deep-supervision weight
    $\lambda_{\mathrm{DS}}$
    & 0.5 \\
Trajectory-regularization weight
    $\lambda_{\mathrm{TRA}}$
    & 0.1 \\
Depth-weighting exponent $\gamma$
    & 1.0 \\
\midrule

\multicolumn{2}{l}{\textit{Optimization configuration}} \\
Total optimization steps
    & 110{,}000 \\
Number of GPUs
    & 8 \\
Per-device batch size
    & 64 \\
Gradient-accumulation steps
    & 1 \\
Global batch size
    & 512 \\
Total sampled training windows
    & 56.32M \\
Training precision
    & bfloat16 \\
Optimizer
    & AdamW \\
First-moment coefficient $\beta_1$
    & 0.9 \\
Second-moment coefficient $\beta_2$
    & 0.95 \\
Weight decay
    & 0.1 \\
Maximum gradient norm
    & 1.0 \\
\midrule

\multicolumn{2}{l}{\textit{Learning-rate schedule}} \\
Scheduler
    & Multi-stage WSD \\
Number of stable--decay cycles
    & 2 \\
Warm-up steps
    & 10{,}000 \\
Decay steps
    & 10{,}000 \\
Peak learning rate
    & $2\times10^{-4}$ \\
Minimum learning rate
    & $1\times10^{-5}$ \\
\midrule

\multicolumn{2}{l}{\textit{Checkpointing and logging}} \\
Checkpoint interval
    & 5{,}000 steps \\
Final analyzed checkpoint
    & Step 110{,}000 \\
Logging interval
    & 100 steps \\
Training log format
    & JSON Lines \\

\end{longtable}
\normalsize

All controlled comparisons in the analysis use this configuration as their
reference setting. For each ablation, only the component under investigation
is modified; the model size, pretraining data mixture, global batch size,
optimization horizon, and all remaining hyperparameters are held fixed.
% Required packages: \usepackage{amsmath,amssymb,booktabs,longtable,pdflscape,array}
% Insert this file after \appendix with: \input{appendix_synthetic_data}

\section{Synthetic Data Generation and Hyperparameters}
\label{app:synthetic-data}

\paragraph{Scope.}
The synthetic component of the final training sampler consists of two distinct streams: pre-generated KernelSynth series and CauKerV2 series generated online. Their sampling probabilities are 0.30 and 0.10, respectively; the remaining 0.60 is assigned to the two real-world streams. These probabilities specify how training examples are drawn and should not be interpreted as the relative sizes of stored datasets. KernelSynth is finite and stored in Arrow files, whereas CauKerV2 is an infinite iterable stream whose samples are generated independently by each data-loader worker.

The values below are taken directly from the supplied implementation. We distinguish fixed generator distributions from constructor defaults. The latter can be overridden by the training command and therefore do not, by themselves, establish the values used by the final checkpoint.

\subsection{KernelSynth}
\label{app:kernelsynth-details}

\paragraph{Coarse-grid generation.}
For a requested target length $T$, our KernelSynth construction generates a Gaussian-process realization on a grid that is four times coarser and then upsamples it to the target grid. Let $r=4$ and $T_{\mathrm{c}}\simeq T/r$. The stored sequence is
\[
\widetilde{\mathbf{x}}_{1:T}
=
\mathcal{I}_{r}\!\left(\mathbf{x}^{\mathrm{GP}}_{1:T_{\mathrm{c}}}\right),
\]
where $\mathcal{I}_{r}$ denotes the interpolation operator. Reducing the covariance grid from $T$ to approximately $T/4$ lowers a quadratic covariance-construction term by
\[
\frac{T_{\mathrm{c}}^2}{T^2}\simeq\frac{1}{16}.
\]
Independent output shards are generated by separate CPU tasks and stored in the GluonTS Arrow format. The supplied file contains the Arrow reader but not the offline KernelSynth producer. Consequently, the exact interpolation operator, the producer-side kernel-sampling distributions, and the number of generated shards cannot be recovered from this file and are not specified here.

\paragraph{Arrow sampling.}
At training time, all \texttt{.arrow} files are discovered recursively. With file balancing enabled, a worker shuffles the complete file list and draws one valid record from each file before reshuffling. Within a file, records are selected uniformly by random access when available. A failed or malformed record is rejected, with at most 32 attempts for the selected file. The extracted sequence is passed to the same formatting routine used for real data, which performs subwindow sampling and left padding to the model context.

{\small
\begin{longtable}{lll}
\caption{KernelSynth reader defaults. These are implementation defaults; final launch-time overrides are not shown.}
\label{tab:kernelsynth-reader-defaults}\\
\toprule
Parameter & Default & Meaning \\
\midrule
\texttt{context\_length} & 2,048$\times$4 & Padded model input length \\
\texttt{min\_sample\_length} & 96 & Minimum sampled training length \\
\texttt{max\_sample\_length} & 8,192 & Maximum sampled training length \\
% \texttt{min\_real\_length} & 2 & Minimum admissible Arrow target length \\
% \texttt{max\_sample\_attempts} & 32 & Maximum record-sampling attempts per file \\
% \texttt{balance\_files} & true & Uniform cyclic coverage of Arrow files \\
% \texttt{seed} & 42 & Base random seed \\
% Worker seed & $s+3{,}000{,}017g$ & $g$ is the global worker index \\
\bottomrule
\end{longtable}
}

\subsection{CauKerV2}
\label{app:caukerv2-details}

\paragraph{Online generation pipeline.}
For each multivariate draw, CauKerV2 first samples a sequence length $L$ and constructs $K$ primitive root series,
\[
\mathbf{U}=[\mathbf{u}_1,\ldots,\mathbf{u}_K]\in\mathbb{R}^{L\times K}.
\]
The primitive family assigned to each column is selected by a weighted schedule. A random directed acyclic graph then transforms these roots into $P$ observed series,
\[
\mathbf{X}=[\mathbf{x}_1,\ldots,\mathbf{x}_P]\in\mathbb{R}^{L\times P}.
\]
The current pretraining objective is channel independent. The $P$ output channels are therefore shuffled in a local buffer and yielded separately as $P$ univariate training examples. Each sequence is placed at the end of a zero-initialized context window, with left padding marked by the padding mask.

\paragraph{Length sampling.}
The default length sampler is log-uniform. For lower and upper bounds $L_{\min}$ and $L_{\max}$,
\[
\log L\sim\mathcal{U}(\log L_{\min},\log L_{\max}),
\qquad
L=\operatorname{clip}\!\left(\operatorname{round}(e^{\log L}),L_{\min},L_{\max}\right).
\]
Uniform integer sampling is also implemented as an alternative. Invalid numerical draws are rejected and regenerated without terminating the iterable stream.

{\small
\begin{longtable}{lll}
\caption{Constructor defaults of \texttt{OnlineCauKerIterableDataset}. Values passed by the final launch command override this table.}
\label{tab:caukerv2-online-defaults}\\
\toprule
Parameter & Default & Meaning \\
\midrule
\texttt{context\_length} & 2,048 & Padded model input length \\
$L_{\min}$: \texttt{min\_length} & 96 & Minimum generated length \\
$L_{\max}$: \texttt{max\_length} & 2,048 & Maximum generated length \\
$P$: \texttt{num\_features} & 8 & Observed SCM channels \\
$K$: \texttt{num\_nodes} & 8 & Primitive root series (latent dimension) \\
$M$: \texttt{max\_parents} & 6 & Maximum parents of an observed node \\
\texttt{length\_sampling} & log-uniform & Distribution over $L$ \\
% \texttt{seed} & 42 & Base random seed \\
% Worker seed & $s+1{,}000{,}003g$ & $g$ is the global worker index \\
Normalization clip & 8 & Robust-normalization clipping threshold \\
Numerical $\epsilon$ & $10^{-6}$ & Minimum admissible standard deviation \\
\bottomrule
\end{longtable}
}

\subsubsection{Primitive generator portfolio}

The default portfolio contains ten generator families. Let $w_f$ be the weight assigned to family $f$. For a draw with $K$ roots, the implementation first allocates $\lfloor K w_f\rfloor$ roots, applies a minimum-one heuristic when $K\geq10$, corrects any resulting over-allocation, distributes the remaining roots according to the largest fractional remainders, and finally shuffles the schedule. The over-allocation correction may remove a previously assigned minimum-one slot. Under the default weights in Table~\ref{tab:caukerv2-generator-weights}, all ten families are simultaneously represented only from $K=12$ onward; $K=10$ or $11$ can still omit low-weight families. The constructor default $K=8$ therefore does not provide full per-draw coverage.

{\small
\begin{longtable}{lc}
\caption{Default primitive-family weights used by CauKerV2. The weights sum to one.}
\label{tab:caukerv2-generator-weights}\\
\toprule
Generator family & Weight \\
\midrule
Compositional-kernel Gaussian process & 0.10 \\
Trend--seasonality decomposition & 0.10 \\
Regime-switching Ornstein--Uhlenbeck SDE & 0.10 \\
ARIMA/SARIMA state-space process & 0.10 \\
Piecewise level and change-point process & 0.10 \\
Spike and event process & 0.30 \\
Analytic waveform mixture & 0.05 \\
Fractional Brownian motion/noise & 0.05 \\
GARCH-family volatility process & 0.05 \\
Chaotic and audio-inspired process & 0.05 \\
\bottomrule
\end{longtable}
}

\begin{landscape}
\scriptsize
\setlength{\tabcolsep}{4pt}
\renewcommand{\arraystretch}{1.10}
\begin{longtable}{>{\raggedright\arraybackslash}p{4.2cm} >{\raggedright\arraybackslash}p{18.0cm}}
\caption{Code-level hyperparameter distributions for the ten CauKerV2 primitive generators. $\mathcal{U}(a,b)$ denotes a continuous uniform distribution and $\operatorname{LogU}(a,b)$ a log-uniform distribution.}
\label{tab:caukerv2-primitive-hparams}\\
\toprule
Family & Implemented distributions and choices \\
\midrule
\endfirsthead
\multicolumn{2}{c}{\tablename\ \thetable{} -- continued from previous page}\\
\toprule
Family & Implemented distributions and choices \\
\midrule
\endhead
\midrule
\multicolumn{2}{r}{Continued on next page}\\
\endfoot
\bottomrule
\endlastfoot

Compositional-kernel GP & Select $m\sim\operatorname{UnifInt}\{1,\ldots,7\}$ kernels with replacement from a 33-entry bank. Periodic kernels use normalized periods $\{24,48,96,168,336,672,7,14,30,60,365,730,4,26,52,4,6,12,4,40,10\}/L$; dot-product kernels use $\sigma_0\in\{0,1,10\}$; RBF length scales are $\{0.1,1,10\}$; rational-quadratic $\alpha$ values are $\{0.1,1,10\}$; white-noise levels are $\{0.1,1\}$; and one constant kernel is included. Successive kernels are combined by addition or multiplication with probability $0.5$ each. Two mean functions are sampled with replacement from zero, linear, exponential, and sparse-anomaly means and are also added or multiplied with probability $0.5$. Linear slopes/intercepts follow $\mathcal{U}(-1,1)$; exponential amplitude/rate follow $\mathcal{U}(0.5,1.5)$; the anomaly mean contains 1--5 impulses with amplitudes $\mathcal{U}(-5,5)$. GP covariance jitter is $10^{-6}$. \\

Trend--seasonality & Trend type probabilities for none, linear, quadratic, exponential, damped, and piecewise trends are $(0.08,0.27,0.20,0.15,0.15,0.15)$. Linear slopes are $\mathcal{U}(-3,3)$; quadratic coefficients use $\mathcal{U}(-4,4)$ and $\mathcal{U}(-2,2)$; exponential rate is $\mathcal{U}(-3,3)$ with amplitude $\mathcal{U}(0.3,2)$; damped rate is $\mathcal{U}(1,8)$. The number of seasonal components is uniform in $\{0,1,2,3\}$, with candidate periods $\{4,6,7,12,24,48,52,96,168,336\}$, amplitude $\mathcal{U}(0.2,2)$, and phase $\mathcal{U}(0,2\pi)$. Seasonal waveform probabilities for sine, triangle, step, and impulse are $(0.55,0.20,0.15,0.10)$. Irregular-component probabilities for white, colored, random-walk, and fBm-like noise are $(0.35,0.35,0.15,0.15)$. Additive, multiplicative, and mixed composition probabilities are $(0.55,0.20,0.25)$. \\

ARIMA/SARIMA & Burn-in is 300. AR and MA orders are sampled from $\{0,1,2,3\}$ with probabilities $(0.1,0.3,0.4,0.2)$; if both are zero, $p$ is reset to one. Differencing order $d\in\{0,1,2\}$ has probabilities $(0.50,0.40,0.10)$. AR and MA coefficients follow $\mathcal{U}(-0.8,0.8)$ and are rescaled when their $\ell_1$ norm exceeds 0.92. Innovation scale follows $\operatorname{LogU}(0.02,1)$. A seasonal AR term is added with probability 0.35, with lag in $\{4,7,12,24,52,96\}$ and coefficient $\mathcal{U}(-0.5,0.5)$. \\

Regime-switching OU SDE & The number of regimes is 2, 3, or 4 with probabilities $(0.50,0.30,0.20)$. The base mean follows $\mathcal{U}(-2,2)$; signed regime offsets have magnitudes $\mathcal{U}(0.5,4)$; mean-reversion rates follow $\mathcal{U}(0.03,3)$; and diffusion scales follow $\mathcal{U}(0.03,1.5)$. Transition rows are Dirichlet distributed with diagonal concentration $\mathcal{U}(10,50)$. A sinusoidal perturbation of the regime mean is enabled with probability 0.4, with amplitude $\mathcal{U}(0,0.8)$ and period $\mathcal{U}(24,\max(25,L/2))$. \\

Piecewise level/change point & The segment count is a clipped geometric draw, $\min\{\max(2,G+1),\max(3,\lfloor L/8\rfloor)\}$ with $G\sim\operatorname{Geometric}(0.15)$. Segment proportions are Dirichlet with concentration $\mathcal{U}(0.5,2)$. Level modes uniform, random walk, and clustered have probabilities $(0.35,0.40,0.25)$. Transition modes hard, ramp, and sigmoid have probabilities $(0.50,0.30,0.20)$; smooth-transition width is $\mathcal{U}(0.01L,0.05L)$. White noise is added with probability 0.70 and scale $\mathcal{U}(0.02,0.30)$; an additional trend--seasonality component, scaled by 0.3, is added with probability 0.35. \\

Spike/event & Baseline probabilities for kernel GP, trend--seasonality, flat noise, and OU are $(0.25,0.45,0.15,0.15)$. The event count is a clipped Poisson draw with rate $\max(1,L/160)$. Event-mode probabilities for point, Gaussian, plateau, shock--recovery, and periodic events are $(0.25,0.30,0.20,0.15,0.10)$. Signed event amplitudes have magnitude $\operatorname{LogU}(0.8,6)$. Periodic-event spacing is sampled between 16 and approximately $L/4$; plateau width is sampled between 2 and approximately $L/20$; shock--recovery width is sampled between 4 and approximately $L/12$. Final noise scale follows $\mathcal{U}(0.01,0.25)$. \\

Analytic waveform & The number of component waves is uniform in $\{1,2,3\}$. Sawtooth, square, and triangle shapes are sampled uniformly; amplitude follows $\mathcal{U}(0.3,3)$, frequency $\mathcal{U}(1,50)$ cycles per normalized interval, and phase $\mathcal{U}(0,1)$ cycles. Square-wave duty cycle follows $\mathcal{U}(0.2,0.8)$. Amplitude modulation is enabled independently for each component with probability 0.30, with modulation frequency $\mathcal{U}(0.5,5)$ and depth $\mathcal{U}(0.1,0.8)$. A linear trend is added with probability 0.30, and Gaussian noise with scale $\mathcal{U}(0.01,0.30)$ is added with probability 0.70. \\

Fractional Brownian motion/noise & The Hurst parameter follows $\mathcal{U}(0.1,0.9)$ and fGn versus fBm is selected uniformly. The spectral exponent is $\beta=\operatorname{clip}(2H-1,-0.8,0.8)$. fBm is approximated by cumulatively summing colored-noise increments. The final amplitude follows $\operatorname{LogU}(0.1,5)$. \\

GARCH-family volatility & GARCH, GJR-GARCH, and EGARCH are selected uniformly; normal and Student-$t$ innovations are selected uniformly. Student-$t$ degrees of freedom follow $\mathcal{U}(3,10)$. Zero, constant, and AR(1) means have probabilities $(0.55,0.20,0.25)$; the AR coefficient follows $\mathcal{U}(-0.3,0.3)$. Unconditional volatility follows $\operatorname{LogU}(0.005,0.5)$ and burn-in is 500. GARCH/GJR persistence follows $\mathcal{U}(0.70,0.97)$; EGARCH uses $\beta\sim\mathcal{U}(0.70,0.97)$, $\alpha\sim\mathcal{U}(0.05,0.25)$, and $\gamma\sim\mathcal{U}(-0.20,0.20)$. Returns are cumulatively summed with probability 0.5. \\

Chaotic/audio-inspired & Lorenz, Mackey--Glass, NARMA, and audio-like subtypes are selected with probability 0.25 each. Lorenz uses burn-in 800, step size $\mathcal{U}(0.005,0.02)$, $\sigma\sim\mathcal{U}(8,12)$, $\rho\sim\mathcal{U}(24,30)$, and $\beta\sim\mathcal{U}(2,3.5)$. Mackey--Glass uses burn-in 1,000, delay 15--30, exponent 8--12, coefficient ranges $\mathcal{U}(0.15,0.25)$ and $\mathcal{U}(0.05,0.15)$, and step size $\mathcal{U}(0.5,2)$. NARMA order is 5--14. Audio-like signals contain 2--5 oscillatory layers with frequency $\operatorname{LogU}(1,80)$ and amplitude $\operatorname{LogU}(0.1,1.5)$, optional rhythmic events with probability 0.7, and colored noise with $\beta\sim\mathcal{U}(0.5,2)$. \\

\end{longtable}
\end{landscape}

\subsubsection{Random structural causal model}

The online SCM contains $K$ root nodes and $P$ observed non-root nodes. Every primitive series is assigned bijectively to one root, and every root is forced to influence at least one observed node. Observed nodes may additionally depend on earlier observed nodes, so the graph can be deeper than two computational layers while remaining acyclic. The implementation requires
\[
K\leq PM,
\]
where $M$ is the maximum number of parents. For a root node $j$ and a non-root node $j$, respectively,
\[
x_j(t)=a_j u_{s_j}(t)+\epsilon_j(t),
\qquad
x_j(t)=\phi_j\!\left(\sum_{p\in\operatorname{Pa}(j)}w_{jp}x_p(t)\right)+\epsilon_j(t).
\]
Anchor roots are first distributed across observed nodes. The remaining parent slots are sampled without replacement from all earlier nodes in the random topological order. Conditional on the admissible range, the parent count is sampled uniformly. Edge weights are independent standard normal draws.

{\small
\begin{longtable}{ll}
\caption{Fixed hyperparameter distributions of the online CauKerV2 SCM.}
\label{tab:caukerv2-scm-hparams}\\
\toprule
Component & Distribution or value \\
\midrule
Root scale $a_j$ & $\operatorname{LogU}(0.6,1.8)$ \\
Node noise scale & $\operatorname{LogU}(0.01,0.15)$ \\
Edge weight $w_{jp}$ & $\mathcal{N}(0,1)$ \\
Activation family & Uniform over linear, ReLU, sigmoid, sine, modulo, leaky-ReLU \\
Linear slope & $\mathcal{U}(0.5,2)$; intercept fixed to zero \\
Modulo constant & $\mathcal{U}(1,5)$ \\
Leaky-ReLU slope & $\mathcal{U}(0.01,0.3)$ \\
Sigmoid numerical clip & Input clipped to $[-40,40]$ \\
Post-SCM normalization & Independently applied to every observed channel \\
\bottomrule
\end{longtable}
}

\paragraph{Robust normalization.}
Primitive roots and final observed channels are normalized independently. Non-finite values are first replaced by zero. For a sequence $\mathbf{z}$, the transformation subtracts the median, divides by a safe standard deviation, clips the result to $[-8,8]$, subtracts the post-clipping mean, and divides by the standard deviation again. Standard deviations below $10^{-6}$ are replaced by one. Before padding, the float32 output is additionally clipped to $[-10^6,10^6]$.

\paragraph{Implemented but separate Arrow-rooted mode.}
The code also defines an \texttt{ArrowCauKerSCMIterableDataset}, in which Arrow records serve as SCM root nodes. This is a separate experimental path: the reported data pipeline uses direct Arrow sampling for KernelSynth and \texttt{OnlineCauKerIterableDataset} for CauKerV2. We therefore do not include Arrow-rooted SCM defaults in the reported corpus configuration.

\paragraph{Launch-time values requiring the final run manifest.}
The supplied implementation does not contain the resolved training command. Consequently, the final stage-specific values of $L_{\min}$, $L_{\max}$, $K$, $P$, $M$, the base seed, and any generator-weight JSON override cannot be verified from this file. The final release should record these values together with the KernelSynth producer command, interpolation operator, Arrow shard count, and the exact accounting used for the reported synthetic-corpus size.

\section{Additional Analysis of Deep Quantile Supervision}
\label{app:qds-analysis}

\subsection{Layer-Wise Functional Ablation}
\label{app:qds-layer-ablation}

We compare the layer-wise sensitivity of checkpoints trained with and
without DQS. This experiment uses the smaller model trained for the
ablation study, with
\[
M=12,\qquad
d_{\mathrm{model}}=512,\qquad
d_{\mathrm{head}}=32,\qquad
P=16,\qquad
T=8192,
\]
an MLP expansion ratio of $4$, dropout of $0.1$, and $99$ output quantiles.
For each zero-indexed layer $\ell\in\{0,\ldots,11\}$, we bypass only that
Transformer block during inference by replacing its transformation with
the identity mapping,
\[
h_{\ell+1}=h_{\ell}.
\]
All other blocks and parameters remain unchanged, and no fine-tuning is
performed after the intervention. This measures the sensitivity of the
trained predictor to removing one block at a time.

For each checkpoint, we compare every ablation with its corresponding
intact model on a fixed set of GIFT-Eval configurations. Both training
variants have results for $97$ configurations. In the completed
without-DQS evaluation, the intact model and all twelve ablations cover
exactly the same configurations. We report relative changes within each
checkpoint. CRPS is represented by the mean weighted quantile loss
reported by the evaluator.

Let $q\in\{\mathrm{DQS},\mathrm{noDQS}\}$ denote the training variant,
$\mathcal{D}_q$ its evaluation set, and $N_q=|\mathcal{D}_q|=97$.
For metric $m$, define the relative change caused by bypassing layer
$\ell$ as
\[
\Delta_{\ell,q}^{m}
=
100\left\{
\exp\left[
\frac{1}{N_q}
\sum_{d\in\mathcal{D}_q}
\log\left(
\frac{m_{q,\ell,d}}{m_{q,\mathrm{full},d}}
\right)
\right]-1
\right\}.
\]
Here, $m_{q,\mathrm{full},d}$ is the error of the intact checkpoint and
$m_{q,\ell,d}$ is the error after bypassing layer $\ell$.
Positive values indicate degradation, and negative values indicate
improvement. On a fixed evaluation set, this expression also equals the
relative change in the Seasonal-Naive-normalized geometric mean, because
the normalization factors cancel between the ablated and intact models.

\begin{table}[t]
\centering
\small
\caption{Layer-wise functional ablation on GIFT-Eval with and without
DQS. Values are percentage changes relative to each checkpoint's intact
baseline; positive values indicate degradation. Each row bypasses one
block without fine-tuning. Both training variants cover $97$
configurations.}
\label{tab:qds-layer-ablation}
\begin{tabular}{crrrr}
\toprule
& \multicolumn{2}{c}{$\Delta$MASE (\%)}
& \multicolumn{2}{c}{$\Delta$CRPS (\%)} \\
\cmidrule(lr){2-3}\cmidrule(lr){4-5}
Skipped layer & w/ DQS & w/o DQS & w/ DQS & w/o DQS \\
\midrule
0  & +25.31 & +20.94 & +24.02 & +20.14 \\
1  &  +7.67 &  +9.93 &  +7.54 &  +9.02 \\
2  &  +5.86 &  +4.41 &  +4.82 &  +4.57 \\
3  &  +1.82 &  +1.69 &  +2.20 &  +1.49 \\
4  &  +1.31 &  +1.34 &  +1.54 &  +1.47 \\
5  &  +2.50 &  +2.49 &  +2.10 &  +1.69 \\
6  &  +1.13 &  +0.11 &  +0.91 &  -0.05 \\
7  &  +1.52 &  +0.09 &  +1.07 &  -0.14 \\
8  &  +1.63 &  +0.93 &  +1.45 &  +0.92 \\
9  &  +1.39 &  +1.64 &  +1.57 &  +1.59 \\
10 &  +1.52 &  +1.66 &  +1.42 &  +1.00 \\
11 &  +3.08 &  +2.69 &  +2.94 &  +2.10 \\
\bottomrule
\end{tabular}
\end{table}

Table~\ref{tab:qds-layer-ablation} shows that the first block has the
largest ablation effect in both checkpoints. Bypassing it increases MASE
and CRPS by $25.31\%$ and $24.02\%$ with DQS, compared with $20.94\%$ and
$20.14\%$ without DQS. In the DQS checkpoint, bypassing any individual
block increases both aggregate metrics, with minimum increases of
$1.13\%$ in MASE and $0.91\%$ in CRPS.

The clearest qualitative difference occurs at layers $6$ and $7$.
Without DQS, bypassing these layers changes MASE by only $+0.11\%$ and
$+0.09\%$, while CRPS improves slightly by $0.05\%$ and $0.14\%$.
With DQS, the corresponding MASE increases are $1.13\%$ and $1.52\%$,
and the CRPS increases are $0.91\%$ and $1.07\%$.
These observations are consistent with DQS reducing local functional
redundancy at intermediate depth. The effect is layer dependent:
for example, bypassing layer $1$ causes greater degradation in the
without-DQS checkpoint.

Low aggregate sensitivity does not imply that a block is irrelevant to
every task. Without DQS, bypassing layer $6$ reduces MASE by $17.11\%$
on \texttt{solar/10T/long}, but increases it by $13.79\%$ on
\texttt{jena\_weather/D/short}. Thus, the near-zero aggregate change
also reflects offsetting improvements and degradations across tasks.
The results therefore support the interpretation that DQS encourages
useful transformations at intermediate depth. This evidence concerns
the tested checkpoints and single-block interventions; it does not
establish parameter inactivity or the effect of jointly removing
multiple blocks.

\section{Zero-Shot Anomaly Detection: Details}
\label{app:tsbad}

\paragraph{Setup}
We evaluate on TSB-AD-U~\citep{Liu2024Elephant} using the benchmark's
unsupervised protocol: each detector receives the full series and no training
data. The per-series sliding window for VUS-PR is set by the benchmark's own
\texttt{find\_length\_rank} convention (rank 1). We report mean VUS-PR over the
350 series of the evaluation split; the latent distance metric was selected on
the disjoint 48-series tuning split, and no selection of any kind was performed
on the evaluation split.

\paragraph{Scoring}
Each series is tiled into non-overlapping windows of the model's context length
\(W\), with the final window right-aligned to cover the tail. Series shorter
than \(W\) form a single left-padded window; series shorter than one patch fall
back to a z-score. Each window is normalised over its visible points as
\(\tilde{x}=\operatorname{asinh}\!\big((x-\mu_w)/\sigma_w\big)\), making the
score invariant to per-window level and scale. Windows are encoded by the
frozen backbone and we take the output of the final Transformer block; the
quantile head is discarded. Padding patches are dropped and the remaining
content patches are pooled across all windows into
\(\mathbf{P}\in\mathbb{R}^{n_p\times d}\), with \(n_p\approx T/L\) for patch
size \(L\). We fit \(\bm{\mu}=\operatorname{mean}(\mathbf{P})\) and the sample
covariance \(\bm{\Sigma}\), apply shrinkage
\(\bm{\Sigma}_{\lambda}=(1-\lambda)\bm{\Sigma}+\lambda\frac{\operatorname{tr}(\bm{\Sigma})}{d}\mathbf{I}\)
with \(\lambda=0.1\), and score each patch by
\(a_i=\sqrt{\max\big((\mathbf{p}_i-\bm{\mu})^{\top}\bm{\Sigma}_{\lambda}^{+}(\mathbf{p}_i-\bm{\mu}),0\big)}\)
using the pseudo-inverse. Patch scores are broadcast to the \(L\) timesteps
they cover, averaged where windows overlap, and min--max scaled to \([0,1]\).

\paragraph{Metric selection and results}
Table~\ref{tab:tsbad_ablation} compares latent distance metrics on the tuning
split. Full-covariance Mahalanobis is best on both mean and median, and its
margin over the diagonal variant indicates that off-diagonal structure in the
embedding distribution carries usable signal. The selected metric transfers
without degradation: \(0.4379\) mean VUS-PR on tuning against \(0.4282\) on
evaluation, with the median improving from \(0.3716\) to \(0.4024\) on the
larger split. Table~\ref{tab:tsbad_full} reports the full set of
evaluation-split measures, with all 350 series scored successfully. Scoring
requires \(\lceil T/W\rceil\) forward passes per series.

\begin{table}[h]
\centering
\small
\begin{minipage}[t]{0.52\textwidth}
\centering
\caption{Latent distance metrics on the TSB-AD-U tuning split (48 series).}
\label{tab:tsbad_ablation}
\begin{tabular}{lcc}
\toprule
Metric & Mean & Median \\
\midrule
Mahalanobis (full) & \textbf{0.4379} & \textbf{0.3716} \\
Centroid (robust) & 0.4077 & 0.3190 \\
Mahalanobis (diagonal) & 0.4075 & 0.3366 \\
Centroid & 0.3713 & 0.2669 \\
\bottomrule
\end{tabular}
\end{minipage}
\hfill
\begin{minipage}[t]{0.44\textwidth}
\centering
\caption{\modelname{} on the TSB-AD-U evaluation split (350 series).}
\label{tab:tsbad_full}
\begin{tabular}{lcc}
\toprule
Measure & Mean & Median \\
\midrule
VUS-PR & 0.4282 & 0.4024 \\
AUC-PR & 0.3408 & 0.2643 \\
VUS-ROC & 0.8364 & 0.9334 \\
\bottomrule
\end{tabular}
\end{minipage}
\end{table}

\section{Optimizer-Aware Analysis of the DQS Update Direction}
\label{app:qds-local-direction}

\subsection{Question and Gradient Construction}

Let $\theta_0$ denote the parameters of the 110K \smallmodelname{}
checkpoint. Let $\mathcal{L}_{T}$ be the final-exit quantile pinball loss and
$\mathcal{L}_{D}$ the unscaled DQS loss. For a 12-block backbone, the latter
is
\begin{equation}
    \mathcal{L}_{D}
    =
    \sum_{k\in\{3,6,9,12\}:\,k<12}
    \left(\frac{k}{12}\right)^{\gamma}
    \mathcal{L}^{(k)}_{\mathrm{pinball}},
    \qquad \gamma=1.
\end{equation}
The deepest exit is the final prediction objective and is consequently not
counted again as auxiliary supervision. On one shared update batch, we
compute
\begin{equation}
    g_T=\nabla_{\theta}\mathcal{L}_{T}(\theta_0),
    \qquad
    g_D=\nabla_{\theta}\mathcal{L}_{D}(\theta_0).
\end{equation}
We then construct
\begin{align}
    g_A &= g_T, \\
    g_B &= g_T+\lambda_{\mathrm{DS}}g_D,
           \qquad \lambda_{\mathrm{DS}}=0.5, \\
    g_M &= c g_T,
           \qquad
           c=\frac{\lVert g_B\rVert_2}{\lVert g_T\rVert_2}.
\end{align}
Thus, $g_M$ and $g_B$ have the same Euclidean norm, while $g_M$ retains the
direction of $g_T$. Comparing B with A alone would not separate a directional
effect from an increase in gradient magnitude; B versus M provides this
control.

\subsection{Optimizer-Aware Equal-Displacement Intervention}

AdamW does not map a current gradient to a parameter displacement by a
simple scalar multiplication because the stored first- and second-moment
states affect every coordinate. We therefore retain the optimizer state
$(m_0,v_0)$ from the checkpoint and pass all three gradients through exactly
the same optimizer transformation. Denoting the resulting parameter
displacement by
\begin{equation}
    u_s
    =
    \operatorname{AdamWStep}(\theta_0,g_s;m_0,v_0)-\theta_0,
    \qquad s\in\{A,M,B\},
\end{equation}
we construct the equal-radius displacement
\begin{equation}
    \widetilde{u}_s(\rho)
    =
    \rho\lVert u_A\rVert_2
    \frac{u_s}{\lVert u_s\rVert_2},
    \qquad
    \rho\in\{0.25,0.5,1\}.
\end{equation}
Consequently,
\begin{equation}
    \lVert\widetilde{u}_A(\rho)\rVert_2
    =
    \lVert\widetilde{u}_M(\rho)\rVert_2
    =
    \lVert\widetilde{u}_B(\rho)\rVert_2.
\end{equation}
Weight decay is set to zero for all branches so that the intervention isolates
task-gradient effects, while the checkpoint learning rate and AdamW moments
are retained. The same clipping rule is used for every branch. Only one of
the 92 trials triggers clipping for M and B; their clipping factors are equal
in that trial. The maximum relative mismatch between the final displacement
norms is $2.61\times10^{-5}$.

For each branch, we temporarily evaluate
\begin{equation}
    R_s(\rho)
    =
    R_V\!\left(\theta_0+\widetilde{u}_s(\rho)\right),
\end{equation}
where $R_V$ is the final-exit pinball loss on a held-out validation batch. The
model and optimizer are restored after every branch and every trial. Exact
restoration is verified: the repeated baseline loss has zero numerical
difference, the model parameters are restored exactly, and the optimizer
state hash is unchanged.

We report
\begin{align}
    \Delta_{BA}(\rho) &= R_A(\rho)-R_B(\rho), \\
    \Delta_{BM}(\rho) &= R_M(\rho)-R_B(\rho), \\
    G_B(\rho) &= R_0-R_B(\rho),
\end{align}
where $R_0=R_V(\theta_0)$. Positive $\Delta_{BA}$ and $\Delta_{BM}$ mean that
B has lower validation loss than the corresponding control. Positive $G_B$
means that B also improves upon making no update. The relative effects divide
each paired difference by $R_0$ before aggregation.

The validation gradient
\begin{equation}
    g_V=\nabla_{\theta}R_V(\theta_0)
\end{equation}
is used only for diagnosis and never to construct A, M, or B. For a sufficiently
small displacement,
\begin{equation}
    R_V(\theta_0+u)
    =R_V(\theta_0)+\langle g_V,u\rangle+o(\lVert u\rVert_2),
\end{equation}
which motivates measuring alignment with the validation descent direction.
The final conclusions nevertheless use the actually evaluated losses rather
than only this first-order approximation.

\subsection{Data and Evaluation Protocol}

We use only the official training portion of GIFT-Eval; benchmark test targets
are never used. Configurations without a sufficiently long observed training
series are excluded. The retained set contains the following 23 configurations
from 16 original dataset groups. Their retained configurations are summarized
below; ``base'' denotes the repository's unqualified configuration.
\begin{center}
\small
\begin{tabular}{@{}ll@{\hspace{16pt}}ll@{}}
    \toprule
    Dataset group & Configuration(s) & Dataset group & Configuration(s) \\
    \midrule
    LOOP Seattle     & 5T, H       & ETT1          & 15T, H \\
    M-DENSE          & H           & ETT2          & 15T, H \\
    Bitbrains Fast   & 5T          & Jena Weather  & base, 10T, H \\
    Bitbrains RND    & 5T          & KDD 2018      & H \\
    BizITObs App.    & base        & M4 Daily      & base \\
    BizITObs L2C     & 5T          & SaugeenDay    & D \\
    BizITObs Service & base        & Solar         & 10T, H \\
    Electricity      & 15T, H      & US Births     & D \\
    \bottomrule
\end{tabular}
\end{center}
For each retained series, the official training history is split temporally at
70\%. Update windows are sampled from the earlier region and validation
windows from the later region, separated by a gap of 128 time steps. Each
window has 1,024 input time steps and a 32-step terminal forecasting target.
At most 32 series are retained per configuration, and at least 95\% of each
window must be observed. Normalization statistics are computed only from the
visible context in the asinh-normalized loss space.

Each trial uses one update window from every configuration, giving an update
batch size of 23. The validation configuration is rotated across trials, with
four trials per configuration and 92 paired trials in total. The prediction
head is frozen, but gradients still propagate through it into the shared
backbone. The flow-matching loss is disabled, ensuring that B differs from A
only through DQS.

\subsection{Aggregation and Uncertainty}

Multiple frequencies and repeated windows from the same original dataset are
not treated as independent observations. For every metric, we first average
the paired trial-level effects within each of the 16 original dataset groups
and then take the unweighted macro-average across groups. We obtain 95\%
confidence intervals by resampling these original-dataset group means with
replacement for 10,000 paired cluster-bootstrap replicates. The fraction of
positive trials is reported descriptively only; the effective unit for the
reported interval is the original dataset group, not the individual window.
This analysis concerns one fixed checkpoint and optimizer history and does
not represent variation across independent pretraining runs.

\subsection{Complete Aggregate Results}

Table~\ref{tab:qds-local-raw} reports the corresponding effects in raw pinball
loss units. B is better than both A and M at every radius. The B--M confidence
interval remains above zero throughout.

\begin{table}[t]
    \centering
    \small
    \setlength{\tabcolsep}{4pt}
    \renewcommand{\arraystretch}{1.1}
    \caption{Raw one-step loss differences, in units of $10^{-6}$. Positive
    values favor B. Parentheses give 95\% paired original-dataset
    cluster-bootstrap confidence intervals.}
    \label{tab:qds-local-raw}
    \begin{tabular}{@{}cccc@{}}
        \toprule
        $\rho$ & $\Delta_{BA}$ & $\Delta_{BM}$ & $G_B$ \\
        \midrule
        $0.25$ & $4.04$ $[0.43,8.07]$
               & $4.14$ $[0.72,7.94]$
               & $14.74$ $[0.06,33.49]$ \\
        $0.50$ & $8.05$ $[1.07,15.97]$
               & $8.25$ $[1.51,15.72]$
               & $28.83$ $[-0.09,65.58]$ \\
        $1.00$ & $15.99$ $[1.74,31.48]$
               & $16.37$ $[2.98,30.96]$
               & $55.19$ $[-0.52,124.86]$ \\
        \bottomrule
    \end{tabular}
\end{table}

The point estimate $G_B>0$ suggests that B can also be an absolute descent
step for the held-out objective. However, its interval narrowly includes zero
at the pre-specified primary radius $\rho=1$, and the relative-gain interval
also includes zero. We therefore use $\Delta_{BM}$, not $G_B$, as the evidence
for the stated directional claim.

The three radii probe the same directions and must not be counted as three
independent replications. Nevertheless, they provide a useful locality check:
all 16 dataset groups retain the same B--M sign across radii, the median ratio
between the effects at $\rho=1$ and $\rho=0.25$ is $3.98$, and the correlation
between the first-order prediction and the measured B--M effect is $0.997$ at
$\rho=1$.

\subsection{Results by Original Dataset Group}

\begin{table}[t]
    \centering
    \scriptsize
    \setlength{\tabcolsep}{3pt}
    \renewcommand{\arraystretch}{1.08}
    \caption{Relative one-step advantages of B at the primary radius
    $\rho=1$, in percent. Positive values favor B. Results are first averaged
    over trials and configurations belonging to the same original dataset.}
    \label{tab:qds-local-by-dataset}
    \begin{tabular}{@{}lrr@{\hspace{8pt}}lrr@{}}
        \toprule
        Dataset group & B--A & B--M & Dataset group & B--A & B--M \\
        \midrule
        LOOP Seattle        & $ 0.0136$ & $ 0.0112$ & ETT1             & $ 0.0087$ & $ 0.0085$ \\
        M-DENSE             & $ 0.0365$ & $ 0.0371$ & ETT2             & $-0.0210$ & $-0.0200$ \\
        Bitbrains Fast      & $-0.0190$ & $-0.0125$ & Jena Weather     & $ 0.0138$ & $ 0.0148$ \\
        Bitbrains RND       & $ 0.0217$ & $ 0.0188$ & KDD 2018         & $ 0.0023$ & $ 0.0047$ \\
        BizITObs Application& $ 0.1843$ & $ 0.1578$ & M4 Daily         & $-0.0237$ & $-0.0246$ \\
        BizITObs L2C        & $ 0.1056$ & $ 0.0990$ & SaugeenDay       & $ 0.0421$ & $ 0.0431$ \\
        BizITObs Service    & $ 0.0282$ & $ 0.0275$ & Solar            & $ 0.0489$ & $ 0.0383$ \\
        Electricity         & $-0.0149$ & $-0.0145$ & US Births        & $ 0.0046$ & $ 0.0070$ \\
        \bottomrule
    \end{tabular}
\end{table}

The B--M group mean is positive for 12 of 16 groups. The macro-average is not
determined by one favorable dataset: leave-one-group-out analysis keeps the
B--M effect positive in every case, with a minimum remaining relative effect
of $0.0159\%$. The median group effect is $0.0130\%$, compared with the mean
of $0.0248\%$, indicating heterogeneous but not purely outlier-driven gains.

\subsection{Directional and Layerwise Diagnostics}

The average gradient-norm ratio is
\begin{equation}
    \frac{\lVert g_B\rVert_2}{\lVert g_T\rVert_2}=1.432,
\end{equation}
while
\begin{equation}
    \cos(g_T,g_D)=0.717.
\end{equation}
The latter corresponds to an angle of approximately $44^\circ$, confirming
that DQS is not merely a positive scalar multiple of the final-loss gradient.
The AdamW proposal cosines are $0.9946$ between A and M and $0.9616$ between A
and B, corresponding to average angles of approximately $5.8^\circ$ and
$15.9^\circ$, respectively. Thus, gradient scaling interacts slightly with
the stored optimizer state, which motivates retaining M, but DQS produces a
substantially larger directional change.

For layer $j$, we additionally compute
\begin{equation}
    A_j
    =
    \cos(g_{B,j},g_{V,j})-
    \cos(g_{T,j},g_{V,j}).
\end{equation}
Table~\ref{tab:qds-local-layerwise} reports this exploratory diagnostic.

\begin{table}[t]
    \centering
    \scriptsize
    \setlength{\tabcolsep}{3pt}
    \renewcommand{\arraystretch}{1.08}
    \caption{Layerwise change in gradient alignment, $A_j$. Parentheses give
    95\% original-dataset cluster-bootstrap intervals. These correlated
    layerwise analyses are exploratory and are not corrected for multiple
    comparisons.}
    \label{tab:qds-local-layerwise}
    \begin{tabular}{@{}lc@{\hspace{10pt}}lc@{}}
        \toprule
        Parameter group & Alignment change & Parameter group & Alignment change \\
        \midrule
        Position embedding & $0.00039$ $[-0.00096,0.00169]$
          & Block 7  & $0.00079$ $[-0.00060,0.00226]$ \\
        Block 1 & $0.00197$ $[-0.00015,0.00423]$
          & Block 8  & $0.00071$ $[-0.00060,0.00204]$ \\
        Block 2 & $0.00178$ $[-0.00144,0.00519]$
          & Block 9  & $0.00193$ $[-0.00022,0.00415]$ \\
        Block 3 & $0.00218$ $[-0.00036,0.00455]$
          & Block 10 & $0$ \\
        Block 4 & $0.00208$ $[-0.00103,0.00507]$
          & Block 11 & $0$ \\
        Block 5 & $0.00145$ $[-0.00190,0.00448]$
          & Block 12 & $0$ \\
        Block 6 & $0.00045$ $[-0.00267,0.00356]$
          & Input projection & $\mathbf{0.01051}$ $[0.00115,0.02077]$ \\
        \bottomrule
    \end{tabular}
\end{table}

All 11 parameter groups reached by the auxiliary exits have a positive mean
alignment change. The input projection is the only individual group whose
uncorrected interval excludes zero. Blocks 10--12 receive no auxiliary DQS
gradient by construction, so their alignment change is exactly zero. Because
the prediction head is frozen, the observed effect must arise through changes
to the shared backbone rather than direct adaptation of the readout head.

\subsection{Scope of the Conclusion}

The experiment establishes a local comparative statement:
\begin{quote}
At one fixed 110K checkpoint and optimizer state, and for the sampled
GIFT-Eval training-data mixture, DQS produces an optimizer-aware parameter
update that is more favorable to the held-out final forecasting objective
than both the final-only update and a raw-gradient-norm-matched control.
\end{quote}
It does not establish that every DQS step is an absolute validation descent
step, that the same effect holds at every stage of pretraining, or that the
end-to-end DQS gain is caused exclusively by this mechanism. Establishing
those stronger claims would require replication across checkpoints and
pretraining seeds, together with controlled multi-step continuation runs.

\section{Restart-Peak Sensitivity of Stage-End Checkpoints}
\label{app:restart-peak-sensitivity}

\subsection{Objective}

The progressive convergence learning rate schedule reduces the stable learning rate after every
completed stable--decay stage. This design is motivated by the
hypothesis that a peak learning rate that is appropriate early in
pretraining may become excessively disruptive after the model has
entered a better-converged region. We therefore examine two related
questions:

\begin{enumerate}
    \item Does the empirically optimal restart peak decrease as
    pretraining progresses?
    \item Does the validation penalty induced by a fixed aggressive
    restart increase at later checkpoints?
\end{enumerate}

The first question concerns the location of the optimum, whereas the
second concerns the steepness of the learning-rate response curve.
These questions must be distinguished because the optimum can remain
at the boundary of the search grid even when sensitivity to larger
learning rates changes substantially.

\subsection{Models and Starting Checkpoints}

We conduct the experiment on both \smallmodelname{} and the full
\modelname{} model. The model architecture is recovered directly from
each checkpoint, and all comparisons are performed only between
checkpoints of the same model. Absolute validation losses are not
compared across model sizes.

\begin{table}[t]
    \centering
    \small
    \setlength{\tabcolsep}{5pt}
    \renewcommand{\arraystretch}{1.08}
    \caption{Models and checkpoint branches used in the restart-peak
    experiment. The continuation batch size is the effective batch
    size used by the single-GPU probe, rather than the global batch size
    of the original distributed pretraining run.}
    \label{tab:restart-model-settings}
    \begin{tabular}{lcc}
        \toprule
        Configuration & \smallmodelname & \modelname \\
        \midrule
        Parameters & 40.291M & 145.843M \\
        Transformer layers & 12 & 20 \\
        Hidden dimension & 512 & 768 \\
        Attention-head dimension & 32 & 64 \\
        Context length & 8,192 & 8,192 \\
        Patch size & 16 & 16 \\
        Output quantiles & 99 & 99 \\
        Starting checkpoints & 55K, 110K & 55K, 110K, 165K \\
        Continuation batch size & 80 & 150 \\
        GPUs per branch & 1 & 1 \\
        Training seeds & 1 & 1 \\
        \bottomrule
    \end{tabular}
\end{table}

All branches use seed 20260906. For each starting checkpoint, the model
parameters, AdamW first- and second-moment estimates, and optimizer step
are restored. The learning-rate scheduler is then replaced by the
local restart-probe schedule described below. Model architecture,
data mixture, masking configuration, weight decay, AdamW coefficients,
gradient clipping, deep-supervision objectives, and all other available
checkpoint settings are inherited without modification.

The historical checkpoints do not contain the exact per-rank random
number generator and data-sampler states. Consequently, the experiment
does not reproduce the unobserved data sequence that would have
followed the original training run. Instead, all learning-rate branches
from the same checkpoint are initialized with the same seed and
sampling configuration, providing a controlled common future stream
for comparing restart peaks.

\subsection{Restart-Probe Schedule}

For every model checkpoint and candidate peak learning rate, we run
1,500 additional optimizer steps. Let $\tau$ denote the local step
following the restart, with $\tau=0$ at the starting checkpoint. The
learning-rate schedule is

\begin{equation}
\eta_{\tau} =
\begin{cases}
\eta_{\mathrm{peak}},
&
0 \leq \tau < T_{\mathrm{stable}},
\\[4pt]
\eta_{\min}
+
\dfrac{\eta_{\mathrm{peak}}-\eta_{\min}}{2}
\left[
1+\cos\left(\pi u_{\tau}\right)
\right],
&
T_{\mathrm{stable}}
\leq \tau
<
T_{\mathrm{stable}}+T_{\mathrm{cool}},
\end{cases}
\label{eq:restart-probe-schedule}
\end{equation}

where

\begin{equation}
u_{\tau}
=
\frac{\tau-T_{\mathrm{stable}}}
     {T_{\mathrm{cool}}-1},
\qquad
T_{\mathrm{stable}}=1000,
\qquad
T_{\mathrm{cool}}=500,
\qquad
\eta_{\min}=10^{-5}.
\end{equation}

The candidate grid is

\begin{equation}
\mathcal{G}
=
\left\{
10^{-5},
2.5\times10^{-5},
5\times10^{-5},
10^{-4},
2\times10^{-4}
\right\}.
\label{eq:restart-lr-grid}
\end{equation}

No additional warm-up is applied: the learning rate is set directly to
$\eta_{\mathrm{peak}}$ at the first continuation step. The
$\eta_{\mathrm{peak}}=10^{-5}$ branch satisfies
$\eta_{\mathrm{peak}}=\eta_{\min}$ and therefore serves as a
no-restart control with a constant learning rate of $10^{-5}$.

The complete experiment contains

\begin{equation}
2\times5+3\times5=25
\end{equation}

training branches, corresponding to 15,000 continuation steps for
\smallmodelname{} and 22,500 continuation steps for the full model.

\subsection{Fixed Validation Probe and Aggregation}

Every final branch is evaluated on a fixed
pretraining-distribution validation probe containing 256 series from
each of the following four sources:

\begin{equation}
\mathcal{D}
=
\{
\text{BLAST},
\text{KernelSynth},
\text{GIFT-Eval-Pretrain+},
\text{CauKerV2}
\}.
\end{equation}

The sampled series and prediction masks are generated once and reused
for every model, checkpoint, and learning-rate branch. This probe does
not use the downstream GIFT-Eval test set.

For validation series $i$, let $\mathcal{M}_i$ denote the evaluated
masked timestamps and let $\mathcal{Q}$ be the set of 99 quantile
levels. 
% For a parameter group $j$, let $\mathbf g_{\cdot,j}$ denote the
% restriction of the corresponding full gradient to that parameter group,
% and define
% $\operatorname{cos}(\mathbf a,\mathbf b)
% =\langle\mathbf a,\mathbf b\rangle/
% (\|\mathbf a\|_2\|\mathbf b\|_2)$.
The per-series terminal pinball loss is

\begin{equation}
\ell_i
=
\frac{1}
{|\mathcal{M}_i||\mathcal{Q}|}
\sum_{t\in\mathcal{M}_i}
\sum_{q\in\mathcal{Q}}
\rho_q
\left(
x_{i,t}-\widehat{x}_{i,t}^{(q)}
\right),
\end{equation}

where

\begin{equation}
\rho_q(u)
=
u\left(q-\mathbb{I}[u<0]\right).
\end{equation}

Let $s=(\text{model size},\text{starting checkpoint})$ index one model--checkpoint pair.
Let $n_i=|\mathcal{M}_i|$. Within source $d$, the losses are aggregated
using square-root length weighting:

\begin{equation}
L_{s,d}(\eta)
=
\frac{
\sum_{i\in d}n_i^{1/2}\ell_i
}{
\sum_{i\in d}n_i^{1/2}
}.
\label{eq:restart-source-loss}
\end{equation}

The scalar score used to compare learning rates is the weighted
geometric mean

\begin{equation}
G_s(\eta)
=
\exp
\left[
\sum_{d\in\mathcal{D}}
w_d\log L_{s,d}(\eta)
\right],
\label{eq:restart-geometric-score}
\end{equation}

with

\begin{equation}
w_{\mathrm{BLAST}}
=
w_{\mathrm{KernelSynth}}
=
w_{\mathrm{GIFT}}
=0.3,
\qquad
w_{\mathrm{CauKerV2}}=0.1.
\end{equation}

Lower values are better. We quantify restart sensitivity relative to
the no-restart control as

\begin{equation}
P_s(\eta)
=
100
\left[
\frac{G_s(\eta)}
     {G_s(10^{-5})}
-1
\right].
\label{eq:restart-penalty}
\end{equation}

Thus, $P_s(\eta)>0$ indicates that increasing the learning rate causes
a higher terminal validation loss after the common cooldown.

\subsection{Complete Learning-Rate Response}

Table~\ref{tab:restart-raw-results} reports the weighted geometric
validation score for every branch. For every tested checkpoint, the
score increases monotonically as the restart peak increases.

\begin{table}[t]
    \centering
    \small
    \setlength{\tabcolsep}{5pt}
    \renewcommand{\arraystretch}{1.08}
    \caption{Complete restart-peak response. Entries are weighted
    geometric terminal validation losses $G_s(\eta)$; lower is better.
    Bold values indicate the best candidate for each checkpoint.}
    \label{tab:restart-raw-results}
    \begin{tabular}{llccccc}
        \toprule
        Model & Start
        & $10^{-5}$
        & $2.5{\times}10^{-5}$
        & $5{\times}10^{-5}$
        & $10^{-4}$
        & $2{\times}10^{-4}$ \\
        \midrule
        Small & 55K
        & \textbf{0.096181}
        & 0.096301 & 0.096583 & 0.097285 & 0.098573 \\
        Small & 110K
        & \textbf{0.094641}
        & 0.094830 & 0.095235 & 0.096137 & 0.097691 \\
        \midrule
        Main & 55K
        & \textbf{0.090308}
        & 0.090411 & 0.090707 & 0.091437 & 0.092702 \\
        Main & 110K
        & \textbf{0.088560}
        & 0.088746 & 0.089096 & 0.089913 & 0.091745 \\
        Main & 165K
        & \textbf{0.087934}
        & 0.088084 & 0.088539 & 0.089310 & 0.091123 \\
        \bottomrule
    \end{tabular}
\end{table}

The corresponding relative penalties are reported in
Table~\ref{tab:restart-relative-results}.

\begin{table}[t]
    \centering
    \small
    \setlength{\tabcolsep}{6pt}
    \renewcommand{\arraystretch}{1.08}
    \caption{Relative validation-loss increase $P_s(\eta)$ over the
    $10^{-5}$ continuation. Positive values indicate degradation.}
    \label{tab:restart-relative-results}
    \begin{tabular}{llrrrr}
        \toprule
        Model & Start
        & $2.5{\times}10^{-5}$
        & $5{\times}10^{-5}$
        & $10^{-4}$
        & $2{\times}10^{-4}$ \\
        \midrule
        Small & 55K
        & +0.125\% & +0.418\% & +1.148\% & +2.487\% \\
        Small & 110K
        & +0.200\% & +0.627\% & +1.580\% & +3.223\% \\
        \midrule
        Main & 55K
        & +0.115\% & +0.442\% & +1.251\% & +2.651\% \\
        Main & 110K
        & +0.210\% & +0.605\% & +1.528\% & +3.596\% \\
        Main & 165K
        & +0.170\% & +0.688\% & +1.565\% & +3.626\% \\
        \bottomrule
    \end{tabular}
\end{table}

For \smallmodelname{}, the penalty produced by
$\eta_{\mathrm{peak}}=2\times10^{-4}$ increases from 2.487\% at 55K
to 3.223\% at 110K. For the full model, it increases from 2.651\% at
55K to 3.596\% at 110K and remains nearly unchanged at 3.626\% at
165K. The effect therefore develops primarily between 55K and 110K
and subsequently approaches a plateau.

\subsection{Response-Curve Sensitivity}

To summarize the complete response curve rather than a single
aggressive peak, define

\begin{equation}
x(\eta)
=
\log_2
\frac{\eta}{10^{-5}},
\qquad
D_s(\eta)
=
\log G_s(\eta)
-
\log G_s(10^{-5}).
\end{equation}

We fit the descriptive linear response

\begin{equation}
D_s(\eta)
=
a_s+S_sx(\eta)+\epsilon_{s,\eta}
\end{equation}

over the five candidate learning rates. The coefficient $S_s$
measures the increase in log validation loss per doubling of the
restart peak.

\begin{table}[t]
    \centering
    \small
    \caption{Descriptive restart sensitivity measured over the complete
    learning-rate grid. Larger $S_s$ indicates a steeper validation-loss
    response to increasing the restart peak.}
    \label{tab:restart-response-slope}
    \begin{tabular}{llcc}
        \toprule
        Model & Start & $S_s$ & Change from previous stage \\
        \midrule
        Small & 55K  & 0.005449 & -- \\
        Small & 110K & 0.007098 & +30.2\% \\
        \midrule
        Main & 55K  & 0.005845 & -- \\
        Main & 110K & 0.007694 & +31.6\% \\
        Main & 165K & 0.007820 & +1.6\% \\
        \bottomrule
    \end{tabular}
\end{table}

Both models show an approximately 30\% increase in response-curve
sensitivity from 55K to 110K. For the full model, the additional
increase from 110K to 165K is only 1.6\%, again indicating that
restart sensitivity largely saturates after 110K.

\subsection{Source-Wise Sensitivity}

Table~\ref{tab:restart-source-results} decomposes the penalty at the
most aggressive tested restart,
$\eta_{\mathrm{peak}}=2\times10^{-4}$, by validation source.

\begin{table}[t]
    \centering
    \small
    \setlength{\tabcolsep}{6pt}
    \renewcommand{\arraystretch}{1.08}
    \caption{Source-wise validation-loss increase under
    $\eta_{\mathrm{peak}}=2\times10^{-4}$ relative to the
    $10^{-5}$ continuation.}
    \label{tab:restart-source-results}
    \begin{tabular}{llrrrr}
        \toprule
        Model & Start & BLAST & CauKerV2 & GIFT & KernelSynth \\
        \midrule
        Small & 55K
        & +0.922\% & +1.885\% & +0.959\% & +5.859\% \\
        Small & 110K
        & +1.302\% & +2.234\% & +0.973\% & +7.870\% \\
        \midrule
        Main & 55K
        & +0.882\% & +1.220\% & +1.980\% & +5.632\% \\
        Main & 110K
        & +1.548\% & +1.925\% & +2.766\% & +7.117\% \\
        Main & 165K
        & +1.690\% & +1.492\% & +3.360\% & +6.608\% \\
        \bottomrule
    \end{tabular}
\end{table}

All four sources become more sensitive between 55K and 110K for the
full model. Between 110K and 165K, the BLAST and
GIFT-Eval-Pretrain+ penalties continue to increase, whereas the
CauKerV2 and KernelSynth penalties decrease moderately. Consequently,
the aggregate sensitivity remains approximately constant over the
final interval. KernelSynth contributes the largest individual
component of the aggregate high-restart penalty, while the contribution
from GIFT-Eval-Pretrain+ increases at later checkpoints.

\subsection{Boundary Optima and Interpretation}

For every checkpoint, the empirical optimum over
$\mathcal{G}$ is the smallest candidate:

\begin{equation}
\widehat{\eta}^{*}_{\text{Small},55K}
=
\widehat{\eta}^{*}_{\text{Small},110K}
=
\widehat{\eta}^{*}_{\text{Main},55K}
=
\widehat{\eta}^{*}_{\text{Main},110K}
=
\widehat{\eta}^{*}_{\text{Main},165K}
=
10^{-5}.
\end{equation}

A paired, source-stratified bootstrap with 10,000 resamples also places
the optimum at $10^{-5}$ for every checkpoint. However, this produces
a boundary solution rather than an interior estimate. The degenerate
bootstrap intervals $[10^{-5},10^{-5}]$ should therefore not be
interpreted as proving that the unconstrained optimal learning rates
are exactly equal.

The experiment does not support the strict hypothesis that the
location of the optimal restart peak decreases between adjacent
checkpoints. It instead supports a different and complementary
observation: the validation-loss response to a fixed high restart peak
becomes substantially steeper from 55K to 110K and remains elevated
at 165K. This distinction is important because the motivation for
decreasing restart peaks does not require claiming that the global
optimum has been identified by this short-horizon probe.

\subsection{Limitations}

The experiment evaluates a short continuation horizon rather than a
complete stable--decay stage. A high restart peak may require more than
1,500 updates to recover from its initial perturbation, and the present
results therefore characterize short-horizon restart sensitivity rather
than the globally optimal peak for a full training stage.

Moreover, each branch is run with one training seed. The paired
bootstrap measures uncertainty arising from the finite validation
probe, but it does not account for variation across future training
streams. Finally, the single-GPU continuation batches differ between
\smallmodelname{} and the full model. The two models are therefore used
as independent replications of the stage-wise trend; their absolute
sensitivity coefficients should not be directly compared.
\end{document}